\documentclass[sn-mathphys-ay,oneside]{sn-jnl}% Math and Physical Sciences Author Year Reference Style
\usepackage[utf8]{inputenc}% UTF-8 encoding support for special characters
\hypersetup{bookmarksopenlevel=3,hypertexnames=false}% Suppress hyperref bookmark level warning; avoid duplicate PDF destinations

\usepackage{graphicx}%
\usepackage{multirow}%
\usepackage{amsmath,amssymb,amsfonts}%
\usepackage{amsthm}%
\usepackage{mathrsfs}%
\usepackage[title]{appendix}%
\usepackage{xcolor}%
\usepackage{textcomp}%
\usepackage{manyfoot}%
\usepackage{booktabs}%
\usepackage{float}
\usepackage{algorithm}%
\usepackage{algorithmicx}%
\usepackage{algpseudocode}%
\usepackage{listings}%
\usepackage{enumitem}
\usepackage{tabularx}
\usepackage{geometry}
\newtheoremstyle{thmstyleone}% name
  {3pt}% Space above
  {3pt}% Space below
  {\itshape}% Body font
  {}% Indent amount
  {\bfseries}% Theorem head font
  {.}% Punctuation after theorem head
  { }% Space after theorem head
  {}% Theorem head spec (can be left empty, meaning 'normal')

\theoremstyle{thmstyleone}%
\newtheoremstyle{thmstyletwo}% name
  {3pt}% Space above
  {3pt}% Space below
  {}% Body font
  {}% Indent amount
  {\bfseries}% Theorem head font
  {.}% Punctuation after theorem head
  { }% Space after theorem head
  {}% Theorem head spec

\theoremstyle{thmstyletwo}%

\newtheoremstyle{thmstylethree}% name
  {3pt}% Space above
  {3pt}% Space below
  {}% Body font
  {}% Indent amount
  {\bfseries}% Theorem head font
  {:}% Punctuation after theorem head
  { }% Space after theorem head
  {}% Theorem head spec

\theoremstyle{thmstylethree}%

\usepackage{acro}

\acsetup{list/display = used, format = \normalfont}

\DeclareAcronym{GAN}{
  short=GAN,
  long=generative adversarial network,
}
\DeclareAcronym{SAM}{
  short=SAM,
  long=Segment Anything Model,
}
\DeclareAcronym{ML}{
  short=ML,
  long=machine learning,
}
\DeclareAcronym{CNN}{
  short=CNN,
  long=convolutional neural network,
}
\DeclareAcronym{Sfm}{
  short=SfM,
  long=Structure-from-motion,
}
\DeclareAcronym{Rv4}{
  short=Rv4,
  long=Norwegian National Road 4,
}
\DeclareAcronym{DFN}{
  short=DFN,
  long=Discrete Fracture Network,
}
\DeclareAcronym{Rhino}{
  short=Rhino,
  long=Rhinoceros 3D,
}

\DeclareAcronym{IoU}{
  short=IoU,
  long=intersection over union,
}

\DeclareAcronym{DPM}{
  short=DPM,
  long=diffusion probabilistic model,
}

\DeclareAcronym{AzureML}{
  short=Azure ML,
  long=Azure Machine Learning,
}

\DeclareAcronym{TP}{
  short=TP,
  long=True positive,
}

\DeclareAcronym{FP}{
  short=FP,
  long=False positive,
}

\DeclareAcronym{FN}{
  short=FN,
  long=False negative,
}

\DeclareAcronym{MoCoA}{
  short=MoCoA,
  long=methods of creation of artifacts,
}

\DeclareAcronym{LLM}{
  short=LLM,
  long=Large Language Model,
}

\acsetup{patch/floats=false} % to ensure acronyms used in tables are counted in the list of acronyms.

\newif\ifreview
\reviewfalse

\ifreview
\usepackage{lineno}
\AtBeginDocument{%
  \linenumbers
  \leftlinenumbers
  \modulolinenumbers[1] % change to [5] if desired
}
\fi

\begin{document}

\title[Article Title]{Automated rock joint trace mapping using a supervised learning model trained on synthetic data generated by parametric modelling}
%! ALTERNATIVE TITLE: Training a generalisable joint trace detection model with few labelled samples


\author*[1,2]{\fnm{Jessica Ka Yi} \sur{Chiu}}\email{jessica.ka.yi.chiu@ngi.no}
\equalcont{These authors contributed equally to this work.}
\author[2]{\fnm{Tom F.} \sur{Hansen}}\email{tom.frode.hansen@ngi.no}
\equalcont{These authors contributed equally to this work.}

\author[2]{\fnm{Eivind M.} \sur{Paulsen}}\email{eivind.magnus.paulsen@ngi.no}
\author[3]{\fnm{Ole J.} \sur{Mengshoel}}\email{ole.j.mengshoel@ntnu.no}


\affil[1]{\orgdiv{Department of Geosciences}, \orgname{Norwegian University of Science and Technology, Norway}, \orgaddress{\street{S.P. Andersens veg 15B}, \city{Trondheim}, \postcode{7031}, \country{Norway}}}

\affil[2]{\orgname{Norwegian Geotechnical Engineering}, \orgaddress{\street{Sandakerveien 140}, \city{Oslo}, \postcode{0484}, \country{Norway}}}

\affil[3]{\orgdiv{Department of Computer Science}, \orgname{Norwegian University of Science and Technology, Norway}, \orgaddress{\street{Sem Sælands vei 9}, \city{Trondheim}, \postcode{7034}, \country{Norway}}}

% \affil[4]{\orgdiv{Department}, \orgname{Organization}, \orgaddress{\street{Street}, \city{City}, \postcode{610101}, \state{State}, \country{Country}}}

\abstract{This paper presents a geology-driven machine learning method for automated rock joint trace mapping from images. The approach combines geological modelling, synthetic data generation, and supervised image segmentation to address limited real data and class imbalance. First, discrete fracture network models are used to generate synthetic jointed rock images at field-relevant scales via parametric modelling, preserving joint persistence, connectivity, and node-type distributions. Second, segmentation models are trained using mixed training and pretraining followed by fine-tuning on real images. The method is tested in box and slope domains using several real datasets. The results show that synthetic data can support supervised joint trace detection when real data are scarce. Mixed training performs well when real labels are consistent (e.g. box-domain), while fine-tuning is more robust when labels are noisy (e.g. slope-domain where labels can be biased, incomplete, and inconsistent). Fully zero-shot prediction from synthetic model remains limited, but useful generalisation is achieved by fine-tuning with a small number of real data. Qualitative analysis shows clearer and more geologically meaningful joint traces than indicated by quantitative metrics alone. The proposed method supports reliable joint mapping and provides a basis for further work on domain adaptation and evaluation.}




\maketitle

\section{Highlights}\label{sec:highlights}
\begin{itemize}
    \item Synthetic rock images facilitate training a core model for detecting joint traces, which can be finetuned with a small real joint dataset.
    \item Training strategies work differently depending on how consistent and clean the real labels are.
    \item Models can produce realistic joint patterns even when standard scores show limited improvement.
    \item Visual assessment is essential because ground truth labels are often incomplete or uncertain.
    \item The method supports reliable joint trace mapping and can reduce the need for extensive manual image labelling.
\end{itemize}

\printacronyms[name=List of Acronyms]

\section{Introduction}\label{sec:introduction}
Digital rock joint mapping mitigates the limitations of conventional in-situ joint surveys, which are often constrained by outcrop accessibility and sampling bias \citep{reid_automated_1997}. It enables full outcrop coverage using images and 3D models and supports consistent extraction of quantitative joint properties for structural characterisation and rock mass stability analysis \citep{battulwar_state---art_2021}.
The application of computer vision in rock engineering has significantly advanced the field of digital rock joint mapping. Traditional semi-automatic, rule-based data processing techniques are increasingly being replaced by automatic \ac{ML} methods, which offer greater adaptability and scalability. Rule-based approaches, while effective to some extent, often require iterative parameter adjustments for each dataset, making them labour-intensive and dataset-specific. In contrast, a well-generalised \ac{ML} model for semantic segmentation could, in theory, handle diverse scenarios with minimal manual intervention. However, achieving such generalisation depends heavily on access to large, high-quality training datasets.

In \ac{ML}, model performance and generalisation are dependent on the size, diversity, and representativeness of the training data. Limited or homogeneous datasets increase the risk of overfitting and lead to models that perform well only under narrowly defined conditions, reducing their robustness in real-world applications. Training \ac{ML} models to identify rock joint traces from images requires manual annotation of thousands of images, a task that is not only labour-intensive but also subjective and error-prone. Rock joint information is inherently scale-dependent; annotating fine cracks within a block may introduce noise when the focus is on identifying block-forming joints. Consequently, obtaining a perfect ground truth for real-world data through manual annotation is virtually impossible, impacting the quality and reliability of \ac{ML} predictions. This highlights the need for datasets that reduces manual effort while improving generalisability.

Current rule-based and \ac{ML}-driven techniques have demonstrated success in extracting joint surfaces from 3D point clouds and images. However, geometric methods often struggle to capture the continuity of joint traces, particularly when limited point cloud resolution results in scattered representations. Supervised ML, leveraging images, point clouds, or a fusion of various data formats, has outperformed rule-based methods in accurately extracting joint trace geometry \citep{Chen2021c, lee_semi-automatic_2022, qiu_autonomous_2025}. 

Despite the advances in rock joint trace segmentation using supervised \ac{ML}, existing models are primarily trained on limited, case-specific datasets, hindering their broader applicability. The Rockbench repository \citep{lato_rock_2013} provides 3D point cloud representations of rock masses, but its use in the literature has been limited to methodological studies, and it does not constitute a curated, large-scale dataset for supervised rock joint mapping. There is no open access, large-scale curated database dedicated to rock joint mapping, as seen in other domains, such as COCO Stuff for general pixel-level annotated images \citep{caesar_coco-stuff_2016}, Cityscapes \citep{cordts_cityscapes_2016}, ADE20K \citep{zhou_scene_2017, zhou_semantic_2019}, and Pascal VOC \citep{everingham_pascal_2009}. Datasets that are more closely related to rock engineering, such as FIND for maps of cracks from bridge and road images \citep{zhou_fused_2022} and Digital Rocks Portal for porous microstructure images \citep{prodanovic_digital_2023}, address fracture-related features but at different scales and for different applications..

%OJM INPUT START

In fact, natural or real-world data, while increasingly available and plentiful, are associated with several limitations.  Specifically, we identify several natural data challenges including these:
(1) Natural data only represent the observed or investigated past.
(2) Natural data may be limited in volume or dimensionality, due to costly data acquisition or processing.  (Here, processing for example refers to adding data labels to examples prior to the use of supervised ML algorithms.)
(3) Natural data may reflect complex or only partially observed phenomena.
(4) Natural data may be highly uncertain or time-varying in nature.

Each of these challenges may represent a serious bottleneck to research and development. To address one or more of the above challenges, researchers consequently create artifacts including synthetic datasets and synthetic fitness functions.  Typically, such artifacts or synthetic datasets are somehow ``inspired'' by natural data or able to create ``natural data'' while having an ability to (fully or mostly) avoid the natural data challenges outlined above.

%OJM INPUT START

Synthetic datasets have shown promise in other domains, such as crack detection in concrete and pavement \citep{kanaeva_road_2021, rill-garcia_syncrack_2022}, plant phenotyping for seed and crop segmentation \citep{toda_training_2020}, planetary rock simulation on the Martian surface \citep{kuang_rock_2021}, microscopic-scale rock mineralogy simulation \citep{ferreira_generation_2022}, etc. However, their application to rock joint mapping remains underexplored. 

\begin{figure}[H]
\centering
\includegraphics[width=\linewidth]{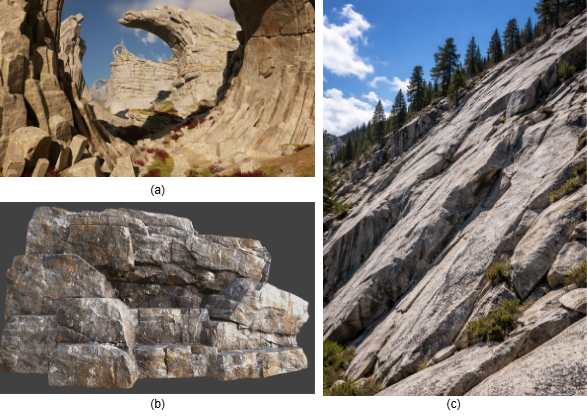}
\caption{Examples of realistic rock outcrop representations available online. (a) The Verdant Realms, a personal 3D environment created by \cite{patscheider_verdant_2024}. (b) Results of a video tutorial example of sculpturing a realistic rock with 3D modelling tools, modified from \cite{olson_quick_2023}. (c) Synthetic rock slope image generated using ChatGPT 5.2 with the prompt “generate an image of a rock slope in granite”.}\label{fig_gaming_rockmass}
\end{figure}

Interestingly, the gaming industry demonstrates effective approaches for creating visually realistic 3D rocky landscapes (Fig.~\ref{fig_gaming_rockmass}a–b), where procedural generation \citep{ebert_texturing_2003} is widely used to algorithmically produce complex geometries and surface textures. While such approaches offer high visual realism and scalability, the generated scenes do not encode explicit physical or geological models, such as joint network geometry or structure. Generative artificial intelligence can likewise produce visually realistic rock images from short text prompts (Fig.~\ref{fig_gaming_rockmass}c). Again, reliable information on rock joint geometry cannot be directly retrieved from these outputs. An automated and parameterised synthetic data generation process provides a unique opportunity to develop scalable and generalised datasets for \ac{ML}-based rock joint mapping.

This study aims to address these rock joint data challenges through a pilot study using synthetic images of jointed rock masses, where joint traces are automatically created in a parameter-controlled and reproducible manner for supervised \ac{ML} training. We propose a novel approach to create scalable and generalisable synthetic datasets by integrating discrete fracture network (DFN) modelling with parametric methods. Focusing on blocky rock outcrops, our work examines the following research questions:\

\begin{enumerate}
    \item How can synthetic jointed rock images be created at the desired scales?
    \item Can \ac{ML} models trained on synthetic data provide reliable predictions for rock joint trace mapping?
    \item How well do these \ac{ML}-models generalise to new scenarios, such as different geology and rock fracturing conditions?
\end{enumerate}

\hspace{0pt}

Our study focuses on joint traces with observable openings (1 mm or wider) on a rock outcrop that can potentially form blocks. These joints are typically resulted from brittle deformation in rocks. Irregularities in the rock related to lithology, such as beddings and foliations that show lineations in intact rock, are not considered. Weakness zones and faults, typically 10 cm or wider, are not included in this work but can be explored in the future as a natural extension.

Previous research and development related to this work are described in Section \ref{sec:related_works}. Datasets and the different methods of preparing each of them, as well as the experimental framework and procedures for \ac{ML} training and validation, are described in Section \ref{sec:methodology}. Results are presented in Section \ref{sec:results} and discussed in Section \ref{sec:discussion}. Our conclusions are drawn in Section \ref{sec:conclusions}. To ensure transparency and reproducibility of our study, we publish our datasets and scripts for generating the synthetic data and \ac{ML} in various open access repositories (GitHub, Azure, and Zenodo), as listed in the supplementary materials. A REFORMS checklist on recommended practice of \ac{ML}-based science following \cite{kapoor_reforms_2024} is also included \ref{sec:appendix_reforms}.

\section{Related works}\label{sec:related_works}

This section is organised into five parts. First, we review the general framework of creation of artifacts. Second, we review the evolution of rock joint trace mapping techniques, highlighting the shift from rule-based methods to \ac{ML}-based approaches. Third, we present recent developments in image-based segmentation using \ac{ML}, including relevant model architectures and annotation strategies. Fourth, we discuss synthetic data as a remedy for annotation quality and dataset availability. Finally, we focus on \ac{DFN} modelling, treated separately due to its unique ability to simulate geologically realistic joint networks. Unlike generic synthetic approaches, \ac{DFN} captures structural logic and joint chronology, making it central to the dataset generation framework proposed in this study.

%OJM INPUT START

\subsection{Methods of creation of artifacts}\label{sec:methods_of_creation}

We identify different classes or \ac{MoCoA} including synthetic data and fitness functions.

We denote the first among the \ac{MoCoA} classes as ``synthesis via construction.''  Here it is up to the researcher or developer to construct meaningful artifacts in the form of synthetic data or fitness functions. Among these artifacts we count synthetic fitness functions used in evolutionary computation.  
Such single- and multi-objective fitness functions can serve as test problems for benchmarking optimization algorithms, including evolutionary algorithms 
\citep{de_jong_analysis_1975,mitchell_royal_1990,goos_constrained_2001,sanchez-diaz_estimating_2024}. 
The fractals of Benoit Mandelbrot can also be counted here \citep{mandelbrot_fractal_1983}. Both fractals and fitness functions are clearly inspired and informed by the natural world, but the exact nature of the mapping from the natural world to the synthetic world is not obvious.

We call the second \ac{MoCoA} class ``synthesis via abstraction.'' In this case, one or more parameters are abstracted from natural data, and these parameters are then used to generate synthetic data. Examples of this includes the use of simulations and synthetic Bayesian networks in artificial intelligence 
\citep{mengshoel_controlled_2006,mengshoel_understanding_2010}.  For example, bipartite Bayesian networks can be used in artificial intelligence to represent, learn, and reason about diagnostic problems.  The two 
partite sets may then respectively consist of diagnoses as root nodes and symptoms as leaf nodes, and 
as the ratio induced by the number of leaf nodes relative to the number of root nodes varies, the treeewidth also appears to vary \citep{mengshoel_understanding_2010}. The notion of treeewidth plays an important role in AI and computer science more broadly, as a measure of computational difficulty   
\citep{bodlaender_efficient_1996,elidan_learning_2008}.

We call the third \ac{MoCoA} class ``synthesis via (data) augmentation.''  In this case one starts with natural data, which are then ``tweaked'' in some fashion to create additional data, which can be fully synthetic data or synthetic data that augments natural data.  Examples of this are often found in computer vision, natural language processing, or applications with simulations. As an example of the latter, there is NASA's ADAPT electrical power system testbed, where signals come partly from a physical electrical power system and partly from an electrical power system  simulation \citep{poll_advanced_2007}. 
ADAPT can then be used to benchmark different methods for diagnosis and prognosis \citep{feldman_empirical_2013}, including methods using Bayesian networks and arithmetic circuits  \citep{mengshoel_probabilistic_2010,feldman_empirical_2013,ricks_diagnosis_2014}.

We call the fourth \ac{MoCoA} class ``synthesis via (data) annotation.''  In this case, data is merely annotated in some form.  Adding labels to examples, in order to enable supervised \ac{ML}, is an example of this. 

%OJM INPUT END

Our methods for creating rock joint datasets as discussed in this work can be classified as the second, third, and fourth \ac{MoCoA} classes. And our use of data augmentation in \ac{ML} training can be considered as the third MoCoA class.

\subsection{Recent developments in automated rock joint trace mapping}\label{sec:evolution_joint_trace_mapping}

The development of automated techniques for rock joint trace mapping has evolved significantly, transitioning from rule-based methods to \ac{ML}-driven approaches. These advancements have emerged parallelly across image-based and point-cloud-based methods, addressing the increasing need for efficient and accurate mapping in diverse engineering project settings.

Rule-based methods employ deterministic algorithms such as edge detection, threshold segmentation, and texture analysis for trace detection. Due to its dependency on user-defined parameters, rule-based methods are typically considered semi-automated. While they excel in simplicity and quick implementation, their dependency on dataset-specific tuning and sensitivity to noise limit their adaptability.

\cite{buyer_applying_2020} developed an edge detection-based method to extract joint traces from digital images, but it struggled with generalising across diverse datasets \citep{chen_machine_2022}. In point cloud-based applications, \cite{guo_geometry-_2019} introduced a geometry- and texture-based approach, demonstrating robustness on smooth rock surfaces but with limited success in regions of low curvature. \cite{guo_automatic_2022} further enhanced this by incorporating colour attributes, addressing limitations in trace detection on low-contrast surfaces. \cite{mehrishal_new_2024} extended the scope of semi-automatic methods by applying photogrammetry-based 3D modelling with curvature analysis and Fuzzy C-Means clustering. Their approach proved effective in complex geological environments, enabling robust trace detection and joint plane orientation calculations. This balance of automation and expert input highlights the potential of semi-automatic methods for complex scenarios.

The shift to \ac{ML}-based methods marked a major advancement in trace mapping by enabling data-driven generalisability and adaptability. Supervised \ac{ML} approaches using 2D images, as highlighted by \cite{Asadi2021}, \cite{Chen2021c}, \cite{lee_semi-automatic_2022}, and \cite{qiu_autonomous_2025} demonstrated superior performance compared to rule-based methods. Similarly, point-cloud-based \ac{ML} methods, such as those developed by \cite{Azhari2021}, leveraged 3D features to enhance joint trace extraction. 

Rule-based methods are computationally efficient and straightforward to implement but struggle with adaptability to varied geological conditions. In contrast, \ac{ML} approaches offer improved generalisability and accuracy but demand significant computational resources and extensive training datasets. This evolution highlights the growing preference for \ac{ML}-based solutions in addressing the complexities of real-world rock mass mapping.

Despite significant methodological progress in automated rock joint trace mapping, existing rule-based and \ac{ML}-based studies are typically developed and evaluated on limited, case-specific datasets with manually annotated joint traces. Such annotations are labour-intensive, subjective, and difficult to control in terms of scale, continuity (persistency), and geometric consistency, which constrains reproducibility and generalisability. These limitations motivate the exploration of alternative data strategies that reduce annotation bias while enabling systematic investigation of \ac{ML} performance.

\subsection{Applications of machine learning to image-based rock joint trace segmentation}\label{sec:ML_joint_trace_mapping}

\ac{ML} has been increasingly adopted for rock joint trace segmentation due to its ability to automate complex segmentation tasks. In this context, Semantic segmentation refers to pixel-wise classification of images and has become a fundamental task in computer vision. Early deep learning approaches were dominated by convolutional encoder–decoder architectures \citep{garcia-garcia_review_2017}, while more recent developments incorporate attention mechanisms, generative modelling, and hybrid frameworks to better capture spatial dependencies \citep{minaee_image_2022, celik_review_2026}.

Image-based rock joint trace segmentation has been inspired by image segmentation in other domains such as crack segmentation in concrete and asphalt, where thin, elongated structural features must be identified from the background material. Models such as DeepLabv3+ \citep{ferrari_encoder-decoder_2018}, UNet++ \citep{zhou_unet_2018}, and CrackSegDiff \citep{jiang_cracksegdiff_2024} have achieved state-of-the-art results in detecting fine cracks and other structural features \citep{bianchi_development_2022, li_highly_2024}.

\Acp{CNN}-based networks have been the dominated models for image segmentation and classification tasks using \ac{ML} \citep{liu_swin_2021}. \acp{CNN} use a hierarchy of convolutional layers to extract spatial features. Each convolution layer in \acp{CNN} apply a fixed-size local receptive field to extract features, making them efficient for local patterns like edges and textures. Methods like FraSegNet, introduced by \cite{Chen2021c}, apply \acp{CNN} using a modified VGG19 backbone \citep{simonyan_very_2014} for segmenting rock joint traces from tunnel face images. \citep{Chen2021c} made visual comparisons of the output images and concluded that FraSegNet performs better than traditional edge detection methods like canny and other \ac{ML} methods including DeepLabv3+, based on fewer noise lines at fracture intersections.

Unlike \acp{CNN}, Vision Transformers use self-attention mechanisms to capture both local and global dependencies within an image. Transformer-based models are primarily developed for natural language processing, using a token as a basic unit that encode a sequence of characters that can range from a single character to a subword and a word \citep{vaswani_attention_2017, devlin_bert_2019}. Vision Transformers \citep{dosovitskiy_image_2020} encode a patch of an image as a token. \cite{qiu_autonomous_2025} introduced a hierarchical transformer-based semantic segmentation framework for image-based fracture trace detection in underground hard-rock environments. Their approach achieved over 10\% better performance than \ac{CNN} networks (U-net and DeepLabv3+) and rule-based edge detection methods under complex lighting and texture conditions. \cite{chen_automatic_2023} adopted a Vision Transformer architecture, Swin-T \citep{liu_swin_2021}, to extract voids from CT-scanned images of coal samples, achieving a higher global accuracy than using U-net and DeepLabv3+ in general. The Swin-T architecture utilises a shift window approach that partitions an image into fixed-size, non-overlapping patches that are shifted between successive layers, enabling a dynamic receptive field for capturing both local and global spatial relationships across the entire image.

Another emerging image segmentation approach is diffusion models, which are a type of generative model that learns to generate data by modelling the process of noise addition and removal to and from data samples, respectively \citep{ho_denoising_2020}. \Ac{DPM}, a subclass of diffusion models that follow a probabilistic framework in the noise addition and removal steps, are primarily used for image synthesis (e.g. image generating tool DALL-E 2) but have recently been adopted for image segmentation. For instance, CrackSegDiff, integrates grayscale and depth images to improve the segmentation of shallow cracks, highlighting the stronger denoising and feature-preserving capabilities by \ac{DPM} than several notable \ac{CNN} and Transformer-based models \citep{jiang_cracksegdiff_2024}. We are not aware of any published work on image-based rock joint segmentation using the \ac{DPM} framework.

Annotation style plays a critical role in the accuracy and robustness of \ac{ML} models. \cite{Chen2021c} explored filled-pixel annotations for estimating joint trace thickness, while \cite{Asadi2021} annotated with straight lines to save computation efforts for \ac{ML} training. Each approach offers distinct advantages: filled-pixel annotations facilitate thickness measurements, whereas lines-based annotations reduce subjectivity and annotation effort. Recent innovations in foundation models that have been pre-trained on a large-scale dataset, such as Meta’s Transformer-based \ac{SAM} \citep{kirillov_segment_2023}, have opened new possibilities for linear crack and trace detection using various annotation styles without any task-specific pre-training. By integrating bounding box prompts with SAM, \cite{rakshitha_crack_2024} achieved high accuracy in identifying fine cracks within complex backgrounds. However, adapting SAM for rock joint trace mapping remains challenging due to the smaller-scale features, the intricate morphology of rock joints, and the homogeneous rock background, which requires domain-specific fine-tuning of the pre-trained models.

Despite the successes of applying \ac{ML} in image-based rock joint trace segmentation, supervised \ac{ML} methods are heavily reliant on high-quality annotated and domain-specific datasets, which remain a critical bottleneck for the generalisability of the \ac{ML}-based approach. To address this gap, the present study investigates the use of parameter-controlled and reproducible synthetic datasets to support supervised \ac{ML} training and to assess model generalisability.

\subsection{The role of synthetic data in machine learning–based image segmentation}\label{sec:synthetic_data_ML}

Recent studies across computer vision show that synthetic data can substantially reduce annotation effort, but its effectiveness depends on how the synthetic–real gap is managed and how synthetic samples are integrated into training. Three recurring strategies are reported: synthetic-only training, mixed synthetic–real training, and synthetic pretraining followed by real fine-tuning. Simulation- and game-engine–based datasets with pixel-accurate labels demonstrate that augmenting limited real data with synthetic samples can approach or, in some cases, exceed real-only performance in segmentation and detection tasks, provided that visual realism and annotation fidelity are sufficient \citep{richter_playing_2016, gaspar_synthetic_2025}. In segmentation and object detection, mixed training and fine-tuning strategies generally outperform synthetic-only approaches, with performance strongly influenced by rendering quality, asset diversity, and domain coverage \citep{cieslak_generating_2024, rasmussen_development_2022, turkcan_boundless_2024}. Systematic benchmarks further indicate that fine-tuning on real data typically outperforms simple mixed training, especially when synthetic data dominates the dataset \citep{wachter_development_2025}.

One notable technique for generating synthetic data is the use of \acp{GAN}. \ac{GAN} is a \ac{ML} framework that consists of two neural networks, a generator that produces data and a discriminator that assesses its authenticity, compete to create highly realistic synthetic data. \acp{GAN} have been successfully applied in various engineering and scientific domains, including structural health monitoring and geological studies. For instance, \cite{bianchi_forecasting_2021} used \acp{GAN} to generate synthetic data for forecasting infrastructure deterioration, demonstrating improved prediction accuracy when trained with such data. Similarly, \cite{ferreira_generation_2022} and \cite{zhao_gan_2023} applied \acp{GAN} to create realistic synthetic rock texture datasets, advancing rock classification and segmentation tasks. However, the effectiveness of \acp{GAN} depends heavily on the availability of large, diverse datasets to generate realistic and generalisable outputs.

In addition to \ac{GAN}-based approaches, deterministic methods based on pre-defined rules have been widely adopted for synthetic data generation. These methods have proven valuable in addressing domain-specific challenges. For example, \cite{rill-garcia_syncrack_2022} developed Syncrack, a tool to generate synthetic images for crack detection in concrete and pavement, showing that accurate synthetic annotations can mitigate the effects of noisy labels and enhance model performance. Similarly, \cite{kanaeva_road_2021} demonstrated the utility of synthetic datasets in road pavement crack detection, where variations in pixel intensity and lighting conditions were effectively captured through synthetic means. Beyond civil engineering, \cite{toda_training_2020} employed synthetic data to train neural networks for crop seed segmentation, achieving high accuracy with reduced annotation effort.

Synthetic datasets offer several advantages, including scalability, annotation efficiency, and reduced training time. They address gaps in real-world data availability and facilitate the exploration of scenarios that are difficult to capture empirically. However, there are ongoing debates about the generalisation capabilities of models trained on synthetic data compared to real-world datasets. Recent analyses have highlighted that synthetic datasets often fail to reproduce the full complexity of real-world visual and physical conditions, which can limit generalisation when models are trained exclusively on synthetic data \cite{kanaeva_road_2021, geng_unmet_2025}. Despite these limitations, synthetic data for rock joint mapping remains largely unexplored. This motivates the current research to investigate the potential of using synthetic datasets for ML-based rock joint trace mapping. By developing a novel deterministic method for generating synthetic jointed rock mass images, this study aims to develop scalable and generalisable solutions for automated rock joint mapping.

\subsection{Discrete fracture networks in rock mass modelling}\label{sec:DFN}
\acp{DFN} are widely used in rock mass modelling to visualise and characterise joint networks at desired mechanical and visual scales. State-of-the-art \ac{DFN} tools such as FracMan, MoFrac, and DFN.lab incorporate advanced capabilities like joint chronology, enabling more realistic simulation of joint network geometries. These tools consider the hierarchical relationships among joints, accounting for features like intersection and termination. The inclusion of joint chronology ensures that DFN models realistically represent rock mass behaviour influenced by sequential fracture formation \citep{erharter_rock_2024, li_critical_2025}.

A notable feature of \ac{DFN} models is their ability to simulate systematic and random joint networks. Unlike random crack generators used in concrete or asphalt e.g. \cite{rill-garcia_syncrack_2022}, rock joint networks exhibit systematic spacing and orientation, occasionally interspersed with random joints. Advanced \ac{DFN} methods accurately model joint terminations, improving the reliability of rock mass characterisation compared to any fully stochastic approach. Metrics such as the Network Connectivity Index (NCI) have emerged to quantitatively assess network properties, capturing fracture intensity, intersection density, and topology in both 2D and 3D domains \citep{sanderson_use_2015, elmo_new_2022, li_critical_2025}.

Synthetic rock mass modelling faces challenges in addressing geological variability and incorporating textural influences from lithology and mineralogy. By integrating \ac{DFN} with parametric techniques, like those developed by \citep{erharter_rock_2024} with finite and folded discontinuities, DFN provides a robust framework for creating synthetic datasets tailored to field-scale joint network simulations.

In this study, a state-of-the-art DFN approach is adopted to model joint networks for synthetic datasets, focusing on capturing joint chronology and connectivity. This methodology facilitates realistic trace mapping, ensuring scalability and adaptability for diverse rock engineering applications.

\section{Methodology}\label{sec:methodology}

\subsection{Datasets and dataset preparation}\label{sec:datasets}

\begin{figure}[H]
\centering
\includegraphics[width=\linewidth]{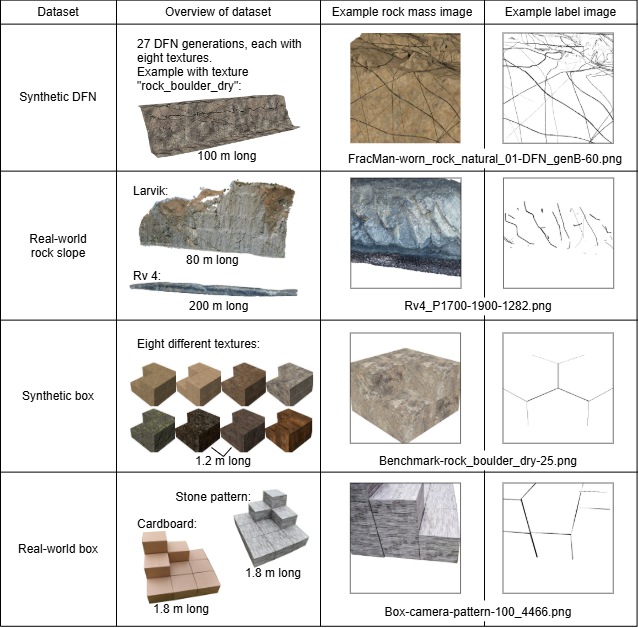}
\caption{Overview and example 2D images of \ac{ML} datasets used in this study.}\label{fig_ML_datasets}
\end{figure}

To investigate the generalisability of \ac{ML} models trained on synthetic data for rock joint detection, we prepare four datasets comprising a total of 29,752 2D images (Table \ref{tab:ML_dataset}). These images constitute the direct inputs to the \ac{ML} models and include both rendered 2D views derived from underlying 3D geometrical models and camera-acquired photographs of real 3D objects in field settings (see Fig.~\ref{fig_ML_datasets}).

The datasets serve complementary roles in training and validating the \ac{ML} models, as described below.

\begin{table}[htbp]
\caption{Overview of 2D images of the \ac{ML} datasets used in this study}\label{tab:ML_dataset}
\begin{tabularx}{\textwidth}{@{}ll>{\raggedright\arraybackslash}X@{}}
\toprule

Dataset              & Number of images & Comments                                                                 \\ 
\midrule
Synthetic \ac{DFN}   & 25584            & Eight textures; 27 \acp{DFN}; generalised rock joint networks                                                   \\ 
Real-world rock slope & 3000             & 1500 each for Larvik and \acs{Rv4} rock cuts                                  \\ 
Synthetic box        & 968              & Five stacked boxes; eight textures                                                               \\ 
Real-world box       & 200              & 100 each for cardbox color and stone pattern                                 \\ 
\midrule
Total                & 29752            & --   \\
\botrule
\end{tabularx}
\end{table}

Each dataset is designed with a specific purpose in the overall ML framework that support controlled experimentation and evaluation of generalisation across synthetic and real domains:

\begin{itemize}
    \item \textbf{Synthetic \ac{DFN} dataset}: This large-scale synthetic dataset is created using FracMan and Rhino/Grasshopper workflows and constitutes the foundational dataset of this study. It covers a wide variety of joint geometries and block shapes, with perfect, automatically created, perfect labels. \textit{Rationale:} Designed to support supervised \ac{ML} training under fully controlled conditions, this dataset enables (i) evaluation of zero-shot prediction on real images, and (ii) investigation of adaptation strategies such as fine-tuning and mixed training with a limited amount of real data. It serves as the primary basis for assessing whether synthetic joint representations can generalise to real-world rock joint mapping tasks.
    
    \item \textbf{Synthetic box dataset}: A minimal and controlled synthetic dataset simulating stacked rock blocks using boxes. Labels are also perfect because they are derived directly from known geometries, eliminating any manual annotation errors. \textit{Rationale:} Provides a simplified training and validation set, useful for early testing and \ac{ML} pipeline debugging.
    
    \item \textbf{Real-world box dataset}: High-resolution camera and drone images of physical cardboard boxes, labelled nearly perfectly to emulate joint traces. \textit{Rationale:} A simplified real-world dataset to evaluate model performance with minimal labelling uncertainty.
    
    \item \textbf{Real-world rock slope dataset}: Includes manually mapped and semi-automatically labelled (thus imperfect) rock joints from two Norwegian road cuts (Larvik and Rv4). \textit{Rationale:} Core evaluation dataset to assess how well models trained on synthetic data generalise to real-world adaptation.

\end{itemize}

\subsubsection{Synthetic discrete fracture network dataset}\label{sec:dataset_synthetic_DFN}

The goal of the synthetic \ac{DFN} dataset preparation process is to create geologically consistent training images of fractured rock cuts with perfect joint
trace labels. The labels are perfect because joint traces are derived directly from known synthetic rock joint geometries without any annotation ambiguity. 

All \acp{DFN} are created for an eastward-facing slope 100 m long and
20 m high, consisting of two 10 m benches at 75° separated by a 1.5 m berm. This fixed slope orientation is adopted purely as a geometric reference to standardise dataset generation and does not constrain the applicability of the resulting images or trained \ac{ML} models. Because joint trace appearance in images is governed by relative orientations between joints and exposed surfaces, rather than absolute geographic orientation, the models trained on this dataset are expected to be generally applicable to rock slopes with arbitrary orientations in real-world settings.

The synthetic \ac{DFN} dataset consists of rendered images (see Section \ref{sec:preprocessing}) of 3D models of synthetic slope surfaces intersected by fracture traces of \acp{DFN}. Structural variability is introduced through controlled
variation in block shape, joint set configuration, and surface texture.

To embed geological variability in the synthetic dataset, block shape is
treated as a parametric variable that controls joint set orientations and
relative intensities. A high-dimensional parametric space describing
parallelepiped geometry is sampled to generate a large population of
candidate block shapes. Each candidate block is classified using commonly
applied rock block shape schemes after \cite{Palmstrom1995} and
\cite{singh_modified_2022}. From this population, 27 representative block
shapes are selected to span the full range of block shape classes and
intermediate geometries.

\begin{figure}[H]
\centering
\includegraphics[width=8cm]{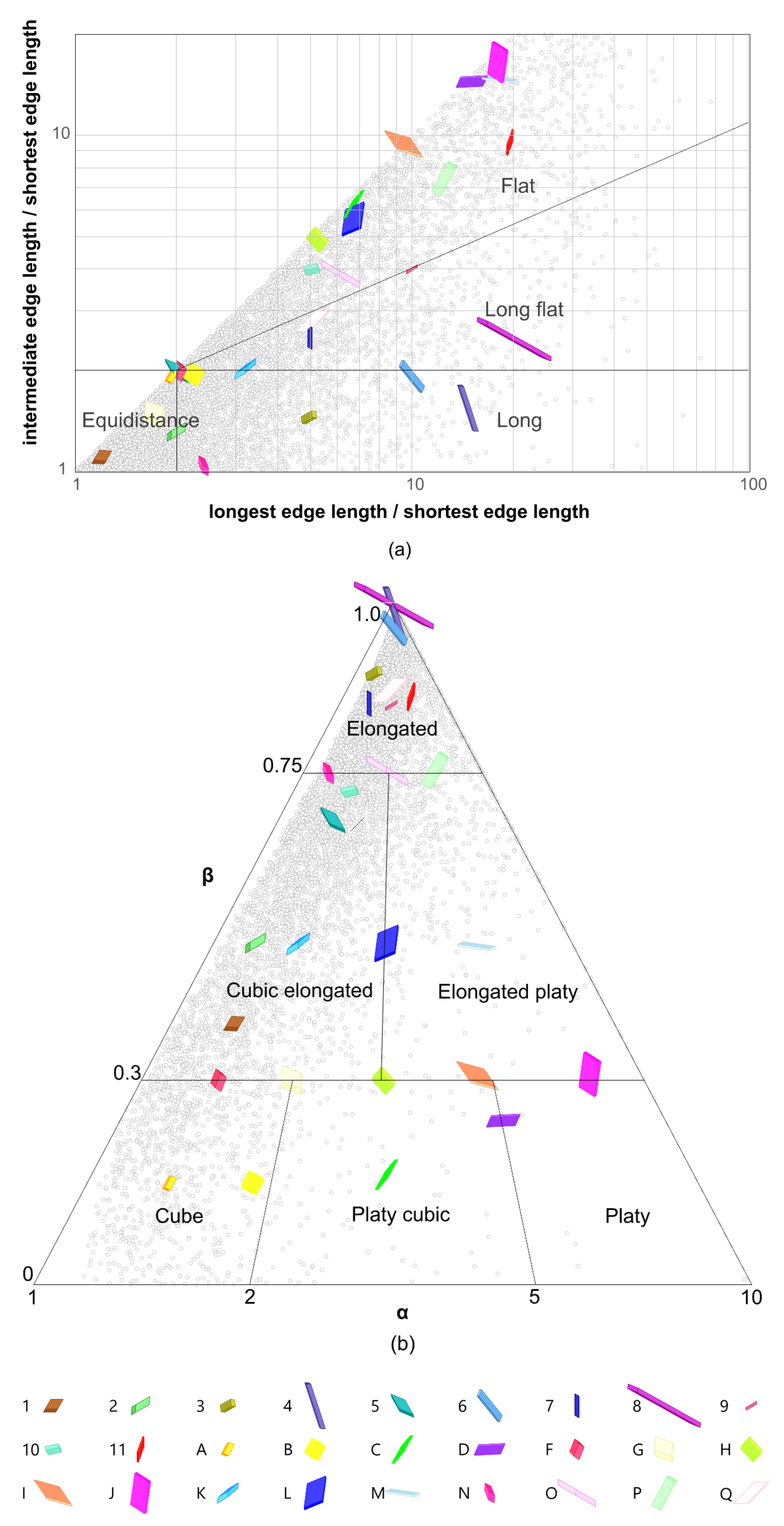}
\caption{Calculated block shape parameters of the 8192 parallelepipeds generated using a parametric study (gray circles) and the selected 27 parallelepipeds (solids in random colours) that span a wide variety of block shape classes based on the classifications (a) after \cite{Palmstrom1995}  and (b) after \cite{singh_modified_2022}. The same parallelepiped is plotted with the same color in both plots.}\label{fig_block-shape}
\end{figure}

The selected block shapes define the orientation of three joint sets and
serve as structural templates for subsequent \ac{DFN} generation. The
distribution of sampled and selected block shapes in the classification
spaces is shown in Fig.~\ref{fig_block-shape}, demonstrating broad and
systematic coverage of block shape variability.

\begin{figure}[H]
\centering
\includegraphics[width=8cm]{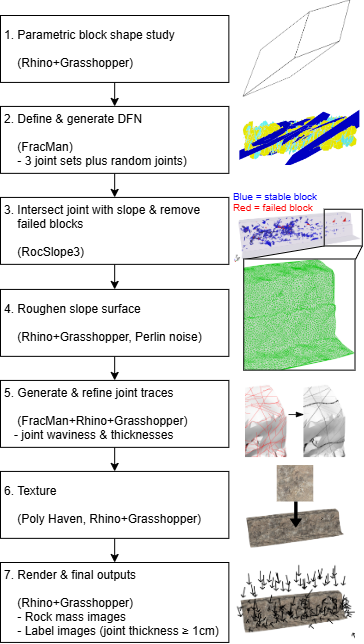}
\caption{Workflow for generating the synthetic \ac{DFN} dataset. Block shapes are created via a parametric study in Grasshopper and used to define joint parameters for \ac{DFN} generation in FracMan. Kinematic analysis in RocSlope3 simulates block removal, and Perlin noise adds surface roughness. Joint traces are extracted, given waviness and thickness, and overlaid on textured slope surfaces. Final images and perfect label masks are rendered for \ac{ML} training.}\label{fig_dataset_synthetic_DFN}
\end{figure}

The workflow for creating the synthetic \ac{DFN} dataset is summarised in Fig.~\ref{fig_dataset_synthetic_DFN}. The synthetic DFN dataset is created using the following procedure:

\begin{algorithmic}[1]
\State Define slope geometry $S$ with fixed orientation and bench layout
\State Generate representative block shapes $\{B_i\}_{i=1}^{27}$
\For{each block shape $B_i$}
    \State Derive joint set orientations $\{\theta_j\}$
    \State Estimate volumetric joint intensity $P_{32,j}$
    \State Generate DFN $\mathcal{F}_i$ with joint sets and random joints
    \State Perform kinematic analysis and remove unstable blocks
    \State Apply surface roughness to slope mesh
    \State Intersect $\mathcal{F}_i$ with slope surface to obtain joint traces
    \State Assign waviness and visual thickness to traces
    \For{each rock texture $T_k$}
        \State Render RGB image and corresponding label mask
    \EndFor
\EndFor
\end{algorithmic}

We assume a constant visual joint thickness $T$ along the joint curves. $T$ is dependent on the length of the intersecting joint curve $L$:

\begin{equation}
T = \frac{L}{\text{100 m}} \times (T_{\text{max}} - T_{\text{min}}) + T_{\text{min}}
\label{eq:thickness}
\end{equation}
where,
\begin{align}
T_{\text{max}} &\text{ is the maximum joint visual thickness}, \nonumber \\
T_{\text{min}} &\text{ is the minimum joint visual thickness}, \nonumber \\
L &\text{ is the length of the intersecting joint curve}.
\end{align}

Details of the parametric formulation, sampling strategy, and software
implementation are provided in ~\ref{sec:appendix_synthetic_dfn}.

\subsubsection{Real-world rock slope dataset}\label{sec:dataset_realworld_slope}

The synthetic rock slope dataset is based on rendered images of coloured point clouds of selected road cuts from the major roads E18 Larvik and \ac{Rv4} Roa-Gran in Norway. The rock joints in the rock cuts are mapped manually using the proprietary software Maptek PointStudio (version 2024). The manual mapping is carried out by selecting points on a single joint surface and/or along its traces in the point cloud. Joint trace labels are generated using the following procedure:

\begin{algorithmic}[1]
\State Manually select points belonging to a joint surface
\State Fit a planar polygon to selected points
\State Apply geometric waviness to the joint surface
\State Intersect joint surface with slope mesh to obtain traces
\State Assign random trace thickness $T \in [1,10]\,$cm
\State Render RGB image and corresponding label mask
\end{algorithmic}

Waviness and thickness are introduced to reflect mapping uncertainty and photographic variability while preserving planar joint geometry. In particular, waviness is applied by displacement of grid points on the planar joint polygon normal to the joint plane with from -1.5 cm to 1.5 cm.

The process of generating the rendered images for \ac{ML} is described in Section \ref{sec:preprocessing}.

The Larvik synthetic rock slope dataset is based on an 80-m long section of a rock cut around 180 m east of the Larvik road tunnel (EU89 UTM32 E560352 N6549559). The rock cut is up to 23 m high and consists of a fractured rock mass with bedrock of Larvikite, a monzonitic igneous rock mainly of alkali felspar. The rock cut was constructed between 2015 and 2017. The Larvik point cloud is derived using the \ac{Sfm} technique with overlapping Google Streetview Images, before its wedge failure in December 2019 \citep{nilsen_landslide_2020}. The Google Streetview images do not cover the top surface of the rock cut and are only limited to a few perspective angles down from both lanes of the E18 highway. This results in zones without sufficient overlap among images or obscured by shadows or other objects. These zones generally lack data points in the point cloud.

The Rv4 synthetic rock slope dataset consists of images from a 200-m long section of a rock cut constructed in 2022. The selected section is southwest-facing and spans between chainage 1700 and 1900. The slope section is located around 130 m east to the northern end of the Garverivegen road in Lunner municipality (EU89 UTM32 E589334 N6685008). The rock cut has an average height of around 8 m and consists of fractured limestone. The Rv4 point cloud is derived by the Sfm technique using drone photos of the newly excavated slope before any rock supports have been installed.

The labelled images are not perfect due to natural surface roughness, variable lighting, occlusion, and subjectivity inherent from the semi-automatic digital rock joint mapping process.

\subsubsection{Synthetic box dataset}\label{sec:dataset_synthetic_box}
The synthetic box dataset consists of rendered images of five boxes stacked into two levels. The boxes have dimensions of 59 cm x 39 cm x 60 cm, equivalent to those in the real-world box dataset (Section \ref{sec:dataset_realworld_box}). Boxes are separated homogeneously with a one-cm opening. The gaps between boxes are shown in the rendered images, simulating rock joints in a jointed rock mass with cubic blocks. We use the same textures as for the synthetic \ac{DFN} dataset (Section \ref{sec:dataset_synthetic_DFN}) to simulate different eight rock types. The textures are mapped to objects directly by using the diffuse channel of the texture image. 

The labels are perfect because gaps between boxes are derived directly from known synthetic box geometries without any annotation ambiguity.

\subsubsection{Real-world box dataset}\label{sec:dataset_realworld_box}

The real-world dataset contains camera and drone images over an outdoor set-up of 13 cardboxes stacked into three levels. The cardboxes have the same dimensions as those modelled in the synthetic box dataset (Section \ref{sec:dataset_synthetic_box}). The surface of the boxes is made of two difference materials -- one with its original cardboard material and kraft paper of the same light brown colour to cover a gap on one of the faces of the boxes; the other is a surface of self-adhesive plastic with a stone pattern. We use the free software Label Studio to manually crop out the background of the images and label the gaps between boxes. 

The labels for this dataset are nearly perfect because gaps between boxes are visually distinct and geometrically simple, allowing consistent manual annotation with minimal subjectivity. However, the manual labelling process can cause slight deviations.

The real-world box dataset serves as a benchmark dataset for rock joint mapping from simulated jointed rock mass images. We contribute to future research by providing the raw images and processed high-resolution point clouds of the real-world boxes in a data repository (Supplementary material \ref{sec:suppl_realworld_box}).  

\subsection{Data preprocessing}\label{sec:preprocessing}
\subsubsection{Rendering for synthetic datasets using Rhino/Grasshopper}\label{sec:rendered_images}
We develop a Grasshopper script to automatically generate multiple rendered images of the object in the viewport in \ac{Rhino} and export them to the desired size and format. The principle of systematically rendering in \ac{Rhino}/Grasshopper in our study is similar to conducting a photogrammetry survey around a rock outcrop based on \ac{Sfm}. In reality, photos are taken with a fixed focal length at certain distances from different locations with respect to the surveyed object. In \ac{Rhino}/Grasshopper, by setting a fixed camera projection and lens length, one only needs to update the camera target and camera location to update the \ac{Rhino} viewport. First, we distribute a specified number of camera target points over the object's surface. We then set the camera location for each camera target point within a certain distance range away from and normal to the object surface. The user specifies the distance range depending on the desired level of details in the images. For all of our synthetic datasets, we randomise the camera distance within the distance range to simulate the varied camera distances in a real photogrammetry survey. Finally, we iterate through each of the camera target-location pairs to update the \ac{Rhino} viewport and export an image for each iteration step. 

Both rock mass and label images are rendered in the same sequence and spatial settings to ensure the image pairs have identical size and spatial alignment. We discover that in some cases, the labels are not shown at a certain rendering angle. This rendering artefact likely needs to be investigated in the graphic setting in \ac{Rhino} in the future.

Some rendered images from the synthetic \ac{DFN} dataset are captured from the side of the slope, showing the backside of the slope surface. Since it will not be possible to capture images from behind the slope surface in reality, these rendered images will not be used for \ac{ML}.

\subsubsection{Standardising image size and format for machine learning}\label{sec:image_preparation}
All the images used for \ac{ML} training and evaluation are of PNG format and 800 $\times$ 800 pixels in dimensions. We use 8-bit images to reduce the memory use during \ac{ML} trainings. The rock mass images used for \ac{ML} have three channels (RGB), whereas all the label images are converted to binary: joint = 0 (black) and non-joint or background = 255 (white) via thresholding.

\subsection{Training and validation}\label{sec:training}

\subsubsection{Experiments}\label{sec:experiments}

The \ac{ML} training is structured to test the hypothesis that a model trained on synthetic datasets, with a controlled proportion of real data, can effectively detect rock joints in real-world images from different sites. We assess both synthetic-to-real generalisation (handling unseen textures or domains) and within-domain adaptation. The scheme consists of ten experiment definitions covering box-like and slope-like scenes. A summary of \ac{ML} training schemes is listed in Table~\ref{tab:ML_training}. Training is divided into three phases for both box-like and slope-like scenes, based on the diversity and domain of the real data used for training compared to those used for testing:

\begin{enumerate}
    \item \textbf{Verification of use of synthetic images:} models are tested on real images from the same broader domain, either box or slope images. This includes the Box and Slope experiments.
    \item \textbf{Within-domain adaptation:} models are tested on images of the seen pattern or lithology, such as cardboard box, stone pattern box, Larvik rock slope, or Rv4 rock slope. This include the Cardboard Box, Pattern Box, Larvik, and Rv4 experiments.
    \item \textbf{Synthetic-to-real generalisation:} models are tested on unseen pattern or lithology. This include the Generalisation Cardboard Box experiment which is trained on real images consisting only of Pattern box; Generalisation Pattern Box only of Cardboard box; Generalisation Larvik only of Rv4 images; and Generalisation Rv4 only of Larvik images.
\end{enumerate}

% Reformulated above; placeholder removed.

As in \cite{wachter_development_2025}, each experiment is executed under two training strategies: Finetune and SimpleMixed. The Finetune strategy adopts a two-stage training: Stage 1 pretrains on synthetic data and validates on real; Stage 2 finetunes exclusively on real data and validates on the same real data. The SimpleMixed strategy trains jointly on synthetic and real data in a single stage, also validating on the same real data as the Finetune strategy. Different proportions of real data are implemented in the experiment sets: 0\%, 10\%, 30\%, 50\%, 70\%, 90\%, and 100\%. The 0\% and 100\% settings are only available with the SimpleMixed strategy and serve as full-synthetic and full-real benchmark experiments, respectively.

\newcolumntype{Y}{>{\raggedright\arraybackslash}X}

\begin{table}[htbp]
\caption{\ac{ML} training experiments and datasets used for training and testing. 
Each experiment is conducted across predefined proportions of real data (0\%, 10\%, 30\%, 50\%, 70\%, 90\%, and 100\%). 
Depending on the training strategy and proportion, training uses either (i) synthetic data only, (ii) real data only, or (iii) a combination of synthetic and real data (e.g., SimpleMixed or Finetune). 
The remaining real data are reserved for testing. 
The total number of images includes both training and test samples. 
Gen.\ = Generalisation.}
\label{tab:ML_training}
\centering
\small
\begin{tabularx}{\textwidth}{@{}lYYY>{\raggedleft\arraybackslash}p{1.2cm}@{}}

    \toprule
    \textbf{Experiment} &
    \textbf{Synthetic training dataset} &
    \textbf{Real training dataset} &
    \textbf{Real test dataset} &
    \textbf{Total images}
 \\
    \midrule
    Box &
    Synthetic box &
    Real-world box &
    Real-world box &
    200 \\

    Pattern box &
    Synthetic box &
    Real-world box (stone pattern) &
    Real-world box (stone pattern) &
    100 \\

    Cardboard box &
    Synthetic box &
    Real-world box (cardboard) &
    Real-world box (cardboard) &
    100 \\

    Gen. pattern box &
    Synthetic box &
    Real-world box (cardboard) &
    Real-world box (stone pattern) &
    100 \\

    Gen. cardboard box &
    Synthetic box &
    Real-world box (stone pattern) &
    Real-world box (cardboard) &
    100 \\

    \midrule
    Slope &
    Synthetic \ac{DFN} &
    Real-world rock slope &
    Real-world rock slope &
    3000 \\

    Larvik &
    Synthetic \ac{DFN} &
    Real-world slope (Larvik) &
    Real-world slope (Larvik) &
    1500 \\

    Rv4 &
    Synthetic \ac{DFN} &
    Real-world slope (Rv4) &
    Real-world slope (Rv4) &
    1500 \\

    Gen. Larvik &
    Synthetic \ac{DFN} &
    Real-world slope (Rv4) &
    Real-world slope (Larvik) &
    1500 \\

    Gen. Rv4 &
    Synthetic \ac{DFN} &
    Real-world slope (Larvik) &
    Real-world slope (Rv4) &
    1500 \\
    \bottomrule
\end{tabularx}
\end{table}

\subsubsection{Evaluation metrics}\label{sec:metrics}
Performance is assessed using standard segmentation metrics and qualitative assessment of predicted masks images.

The quantitative metrics are based on the statistics of the classification of each pixel in the predicted mask image:
\begin{enumerate}
    \item \ac{TP} – pixels correctly identified as joint traces, appearing in both the predicted mask and the label mask;
    \item \ac{FP} – pixels incorrectly predicted as joint traces, present in the predicted mask but not in the label mask;
    \item \ac{FN} – pixels missed by the model, present in the label mask but absent in the predicted mask.
\end{enumerate}

Using these classifications, we use four evaluation metrics that characterise both the accuracy and robustness of the model: 

\vspace{0.5\baselineskip}

\begin{equation}
\mathrm{IoU} = \frac{TP}{TP + FP + FN}
\label{eq:iou}
\end{equation}

\vspace{0.5\baselineskip}

\begin{equation}
\mathrm{Dice} = \frac{2TP}{2TP + FP + FN}
\label{eq:dice}
\end{equation}

\vspace{0.5\baselineskip}

\begin{equation}
\mathrm{Precision} = \frac{TP}{TP + FP}
\label{eq:precision}
\end{equation}

\vspace{0.5\baselineskip}

\begin{equation}
\mathrm{Recall} = \frac{TP}{TP + FN}
\label{eq:recall}
\end{equation}

\Ac{IoU} and the Dice Coefficient, which measure the overlap between the predicted and true positive pixels, are used to evaluate the accuracy of the model in detecting rock joints in the image. \ac{IoU} particularly effective for assessing spatial alignment between predicted and actual joint traces, while the Dice Coefficient is more sensitive to the presence of small or thin features, making it suitable for assessing how well the model captures the fine-grained joint traces. Precision indicates the proportion of predicted joint traces that are correct, whereas recall reflects the proportion of actual joint traces that were successfully detected by the model.

The evaluation considers classification performance for the overall segmentation, joint traces (foreground class) and non-joint traces (background class).

In the present datasets, joint traces constitute a small minority, whereas background pixels dominate the images. 
As a result, performance metrics computed explicitly for the background class are inherently insensitive, as high background scores can be obtained even when joint detection performance is poor. 
Reporting background-based metrics would therefore provide limited insight into the model’s ability to identify structurally relevant features.
For this reason, model evaluations in Section \ref{sec:quantitative_analysis} is based on metrics computed for the joint trace (foreground) class, while the background class is implicitly accounted for through false positive terms.

In addition to quantitative metrics, we perform a visual inspection of the segmentation results of selected epochs per strategy: SimpleMixed includes, if available, epochs 5, 10, 15, 20, and final; Finetune includes epochs 5 and 10 in Stage~1, and first and fifth epochs in Stage~2, plus the best epoch. We rate final/best epochs using four criteria on a 1--5 scale: 

\begin{enumerate}
    \item Geological recognisability: How realistic and geologically plausible do the predicted joints appear?
    \item Joint persistence: Are continuous joints properly connected without gaps or interruptions?
    \item Boundary localisation \& thickness: Are joint boundaries precise and is the thickness appropriate?
    \item False positives/ noise: How much spurious segmentation or noise is present? (1=very noisy, 5=very clean)
\end{enumerate}

\subsubsection{Machine learning training} \label{sec:ML_training}

All models are trained using supervised learning, where the input is a 2D RGB image and the output is a binary mask classifying each pixel as either a joint trace (positive) or background (negative). \ac{ML} training is implemented in Python version 3.12, using core functionality from PyTorch and the Segmentation-Models \citep{iakubovskii_segmentation_2019} library for model architectures and training utilities. \acp{LLM} were used as a supporting tool during software development. Specifically, \acp{LLM} were used to assist with code generation and debugging of machine learning training, evaluation, and data processing scripts.

Two \ac{CNN} architectures, U-net \citep{ronneberger_u-net_2015} and DeepLabv3+ \citep{ferrari_encoder-decoder_2018} are run for comparisons. They are selected due to their proven performance in related crack segmentation (e.g. \citealp{bianchi_development_2022, Dais2021, lee_semi-automatic_2022}. Transfer learning is implemented to accelerate convergence and improve generalisation by using pre-trained encoder backbones from ImageNet in both models.

Training uses a standard 90/10 split for training and validation, ensuring that test sets are unseen and held out during model development. Across both Finetune and SimpleMixed strategies, all models are trained with the Dice loss function, chosen for its sensitivity to class imbalance and ability to focus learning on small, thin structures such as rock joint traces. A default prediction threshold of 0.5 is used for the minority class.

Validation/early stopping are monitored on real data only, using the Dice score on the minority joint class. We use Adam with initial learning rate 0.001 and a ReduceLROnPlateau scheduler (mode=max, factor=0.5, patience=5), yielding halving steps such as 0.001 → 0.0005 → 0.00025 → 0.000125 when the Dice score plateaus. To mitigate overfitting during Stage 2 on limited real data, we cap the carried learning rate at 3.125e-5 (approximately 1/32 of the initial rate); if the current rate is already lower, it remains unchanged. Patience for early stopping is set to ten epochs, reduced to five in Stage 2 for Finetune to avoid overfitting. Batch size is eight. Each experiment runs for a maximum of 100 epochs.

Data augmentation (implemented using the Python library Transforms in PyTorch) is applied only to the training split to increase diversity and reduce overfitting: random crop (65–100\% area) resized to 768×768; horizontal and vertical flips (each 50\% probability); rotations of ±15° with white fill; random affine transforms (±10\% translation, 0.8–1.2× scale, ±10° shear); random perspective (distortion scale 0.2, 50\% probability); colour jitter (brightness ±40\%, contrast ±40\%, saturation ±30\%, hue ±15\%); and Gaussian blur (3×3 kernel, $\sigma$=0.1–1.0).

\subsubsection{Computing infrastructure}\label{sec:infrastructure}
Azure Machine Learning (Azure ML) is used for training and validation. Azure ML provides access to high-performance GPUs for faster model training, parallel execution, automatic logging of the training process, experiment tracking, and dataset storage with versioning. In addition to supporting research experiments, Azure ML is a production-ready platform, allowing trained models to be stored, reproduced, and, if required, deployed in production within a consistent computing environment. The same training pipelines can be reused for continuous retraining as new field data become available.

For all experiments, we used virtual machines from the Standard\_NC6s\_v3 series: six virtual CPUs, 112 GiB memory, and an NVIDIA Tesla V100 GPU with 16 GiB RAM. We use two parallel subprocesses for data loading to minimise transfer delays during storage and training.

\section{Results}\label{sec:results}

\subsection{Degree of generalisation of the synthetic discrete fracture network dataset}\label{sec:Gen_training_datasets}
\begin{figure}[H]
\centering
\includegraphics[width=\linewidth]{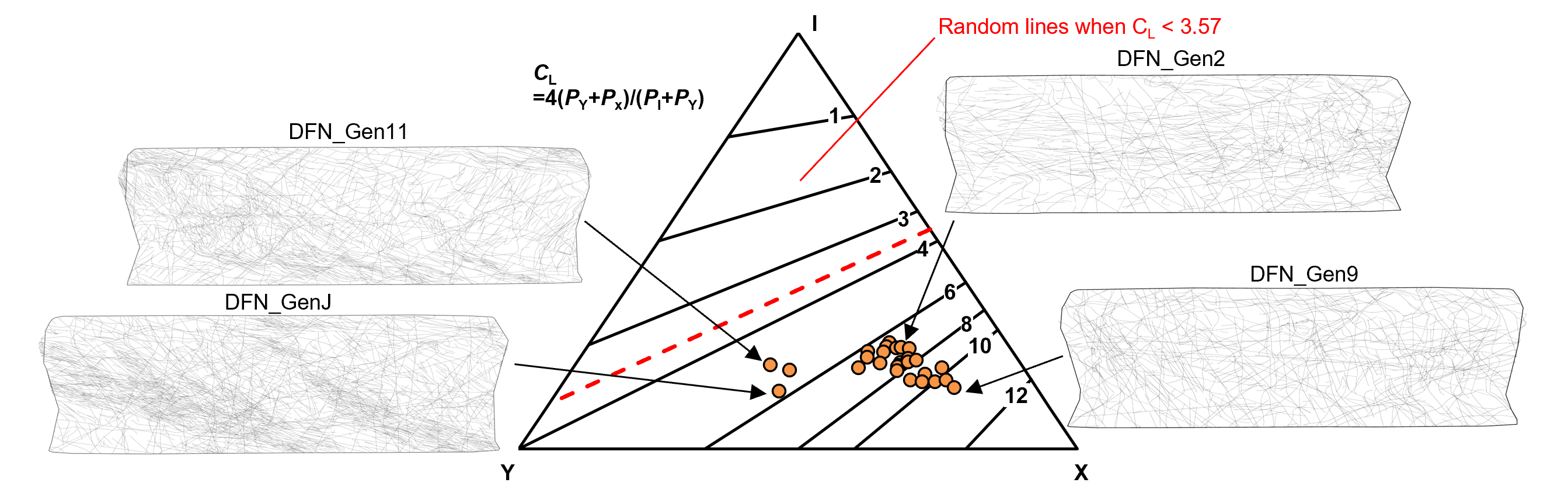}
\caption{Triangular plot of the proportion of I-, X- and Y-nodes for the joint trace networks in all the 27 \ac{DFN} models (after \cite{manzocchi_connectivity_2002, sanderson_use_2015}). ${C_{L}}$ represents the average number of connections per line. Examples of \ac{DFN} models in two clusters of joint network connectivities are shown, together with the edge boundary of the slope face.}\label{fig_network}
\end{figure}

Analysis of joint trace network topology shows a high proportion of Y-nodes across the synthetic \acp{DFN}, indicating that the generated networks do not resemble purely stochastic systems. This is supported by connectivity index ($C_{L}$) values remaining above 3.75, below which being typically associated with random line simulations \citep{balberg_excluded_1984, sanderson_use_2015} (Fig.~\ref{fig_network}). As shown in Fig.~\ref{fig_network}, the networks cluster into two distinct types: one with bands of closely spaced fractures and another with more uniformly distributed fracture spacing.

\begin{figure}[H]
\centering
\includegraphics[width=0.8\linewidth]{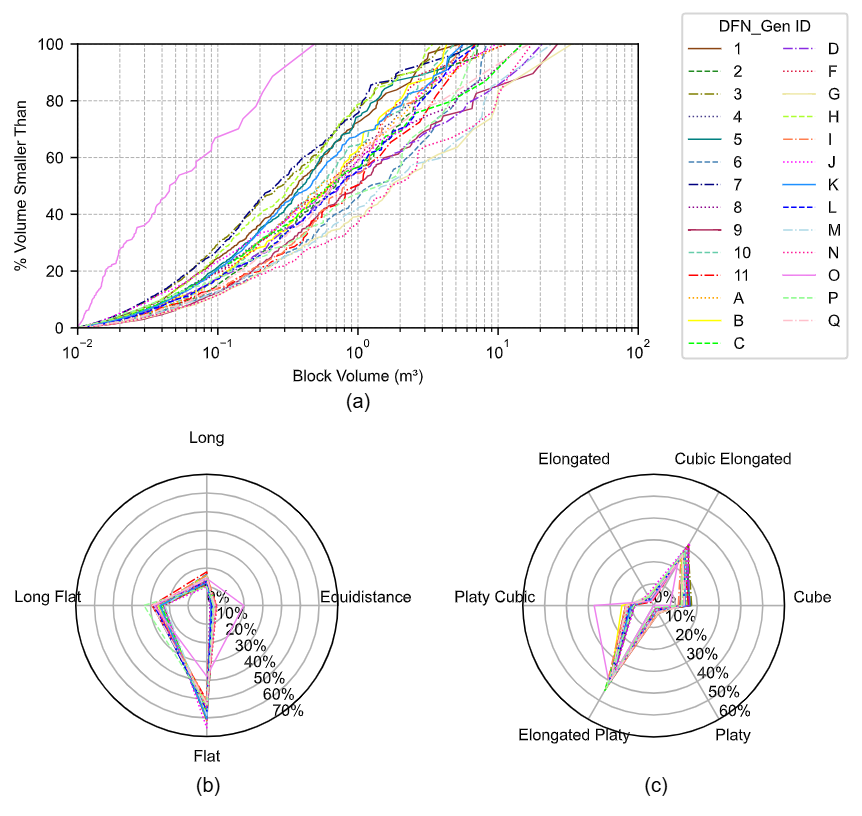}
\caption{Distributions of (a) block volume and (b-c) shape for each \ac{DFN}. In (a), the horizontal axis shows block volume ($m^3$) on a logarithmic scale, and the vertical axis shows the cumulative percentage of blocks with volume smaller than the corresponding value. In (b) and (c), radial axes indicate the percentage share of blocks in each block shape class, while angular axes represent block shape classification schemes by \cite{Palmstrom1995} and (c) by \cite{singh_modified_2022}.}\label{fig_block_volume_shape_distribution}
\end{figure}

Block volume and shape distributions for the 27 \acp{DFN} are presented in Fig.~\ref{fig_block_volume_shape_distribution}. Most models exhibit a relatively even spread across block sizes, with dominant block shapes classified as flat to long-flat according to \cite{Palmstrom1995}, and elongated platy to cubic elongated according to \cite{singh_modified_2022}. A notable exception is \ac{DFN} Gen O, which is characterised by smaller block volumes and a higher proportion of equidimensional shapes by \cite{Palmstrom1995} and platy cubic shapes by \cite{singh_modified_2022}.

\subsection{Machine learning results}\label{sec:ML_results}

A total of 240 experiments were evaluated using quantitative overlap metrics and a complementary qualitative assessment. Fig.~\ref {fig_all_experiments_grid} to Fig.~\ref {fig_dice_vs_quality_grid} present the trends and exemplars. The results are organised to emphasise similarities and differences between training strategies (Finetune vs. SimpleMixed), network architectures (U-net vs. DeepLabv3+), and experimental domains (Box vs. Slope, non-generalisation vs. generalisation, and specific sub-domains).

\subsubsection{Quantitative analysis}\label{sec:quantitative_analysis}

\begin{figure}[H]
\centering
\includegraphics[width=\linewidth]{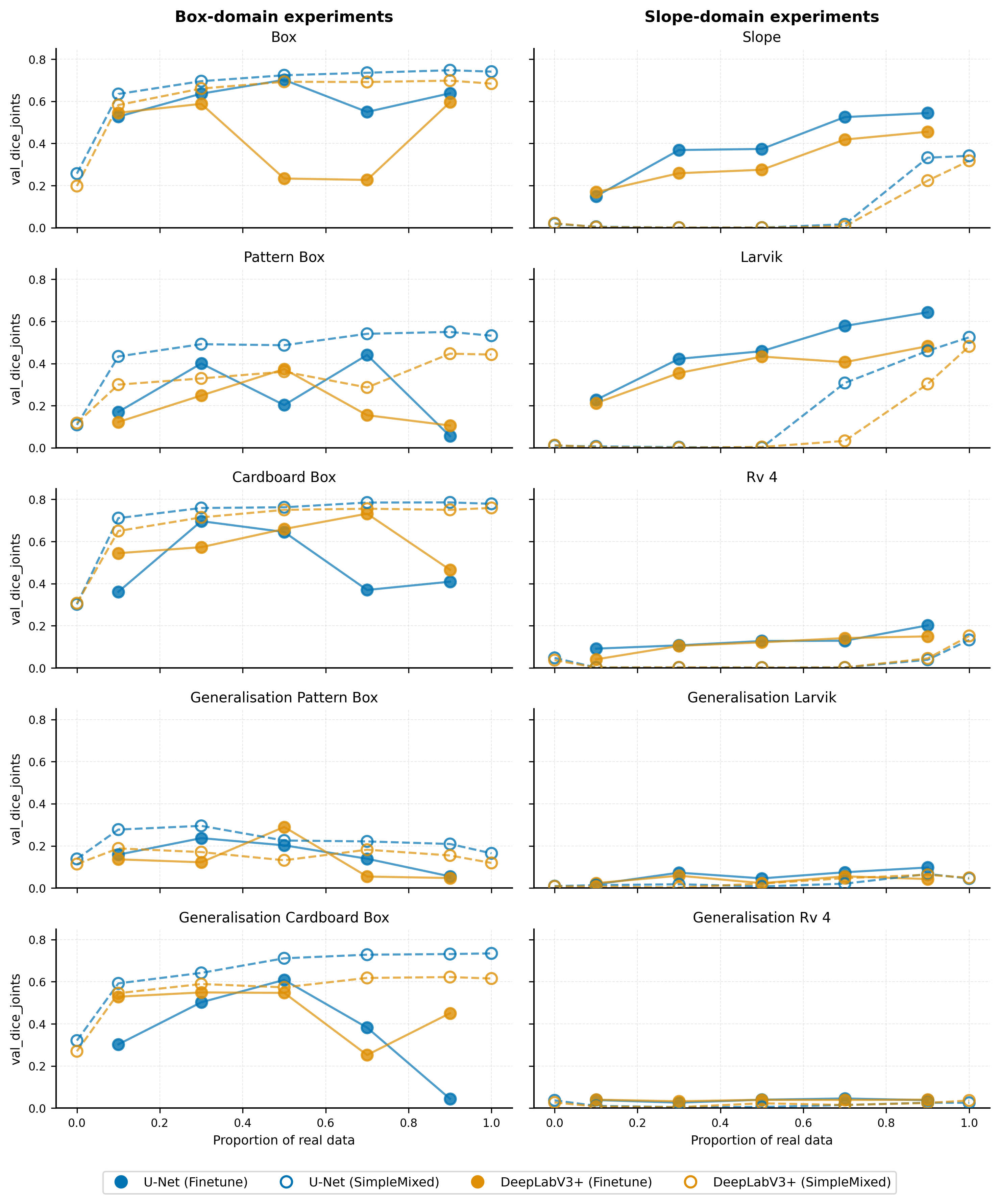}
\caption{Validation Dice (joints) across strategies (Finetune and SimpleMixed) and real-data proportions for U-net and DeepLabv3+.}
\label{fig_all_experiments_grid}
\end{figure}

\paragraph{Box-domain experiments:}
Across Box, Pattern Box, and Cardboard Box experiments, SimpleMixed generally performs on par with or better than Finetune in terms of validation Dice (joints) (Fig.~\ref{fig_all_experiments_grid}). Improvements are typically realised at low proportions of real data (0--10\%), after which performance plateaus or oscillates rather than increasing monotonically. In several Box variants, Dice decreases toward 100\% real data, indicating that increased real-data proportion does not necessarily translate to improved overlap once the synthetic structural prior is diluted.

Among Box sub-domains, Cardboard Box consistently achieves higher Dice than Pattern Box across strategies and architectures, reflecting clearer geometry and lower texture ambiguity. Generalisation Box experiments exhibit reduced Dice overall, with performance often peaking at intermediate real-data proportions (approximately 30--50\%) before declining, particularly for U-net.

\paragraph{Slope-domain experiments:}
In contrast, Slope-domain experiments (Slope, Larvik, and Rv4) benefit more clearly from the Finetune strategy. Validation Dice (joints) generally increases with higher real-data proportions, especially between 10\% and 50\% (Fig.~\ref{fig_all_experiments_grid}). SimpleMixed performs poorly at low real-data proportions in slope settings and only shows notable improvements at high proportions (70--100\%).

Larvik consistently outperforms Rv4 across strategies and architectures. Generalisation to unseen slopes yields uniformly low Dice values (typically $<0.1$), most pronounced for Generalisation Rv4, indicating a substantial domain shift between training and test images.

\paragraph{Architectural comparison:}
Overall, U-net and DeepLabv3+ show comparable quantitative performance (Fig.~\ref{fig_all_experiments_grid}). Under the SimpleMixed strategy, U-net consistently achieves higher validation Dice (joints) than DeepLabv3+ across most domains. In contrast, architectural differences are minimal in Rv4, Generalisation Rv4, and Generalisation Larvik, where both models perform similarly and at low Dice levels.

\subsubsection{Qualitative analysis}\label{sec:qualitative_analysis}

\begin{figure}[H]
\centering
\includegraphics[width=\linewidth]{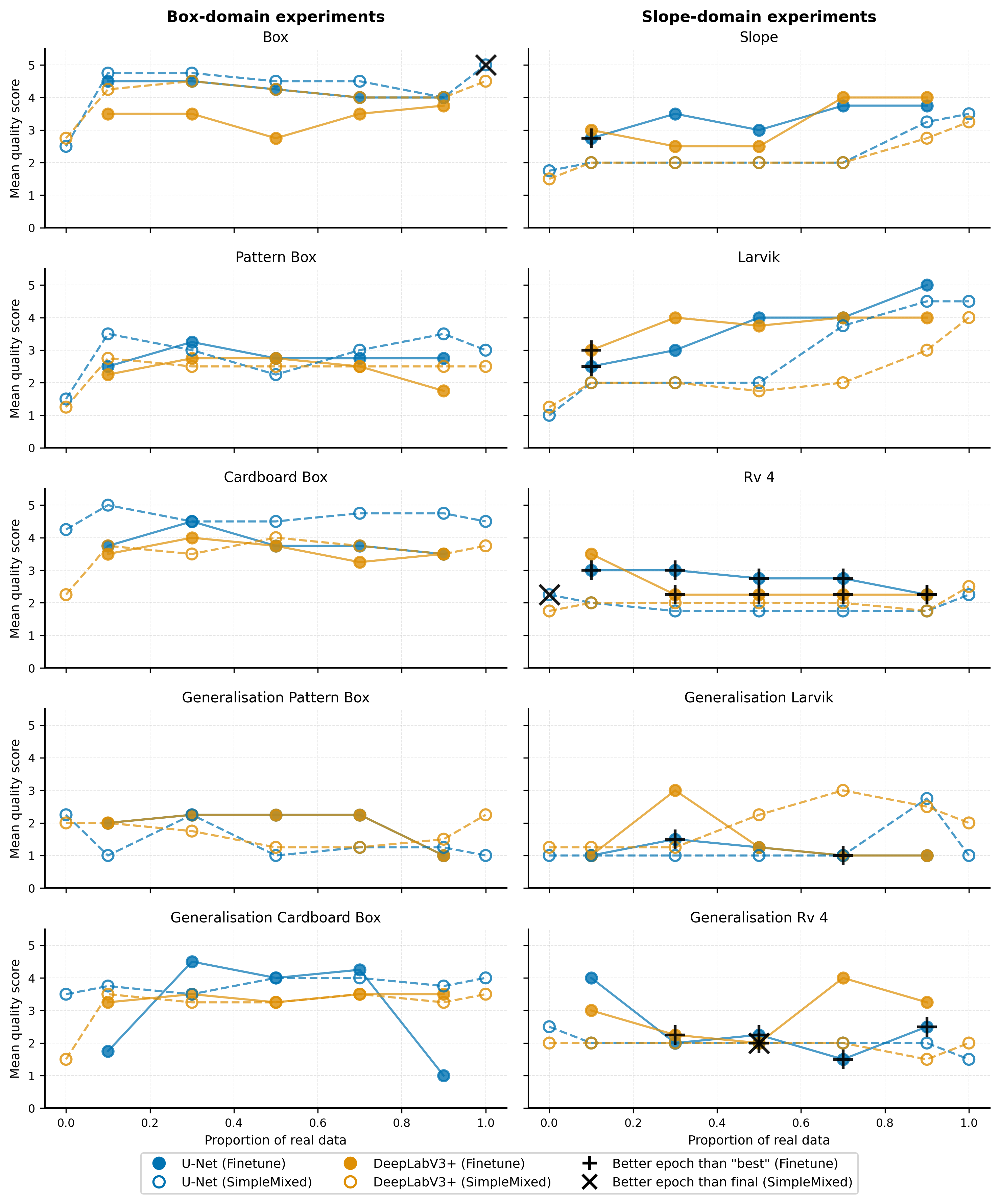}
\caption{Mean quality scores (1--5) for strategies and data proportions, averaged over four criteria.}
\label{fig_qualitative_ratings_grid}
\end{figure}

Qualitative assessment, which is based on the mean quality score from the abovementioned criteria, provide a complementary perspective focused on geological plausibility and engineering usability rather than pixel-level overlap (Fig.~\ref{fig_qualitative_ratings_grid}).

\begin{figure}[H]
\centering
\includegraphics[width=\linewidth]{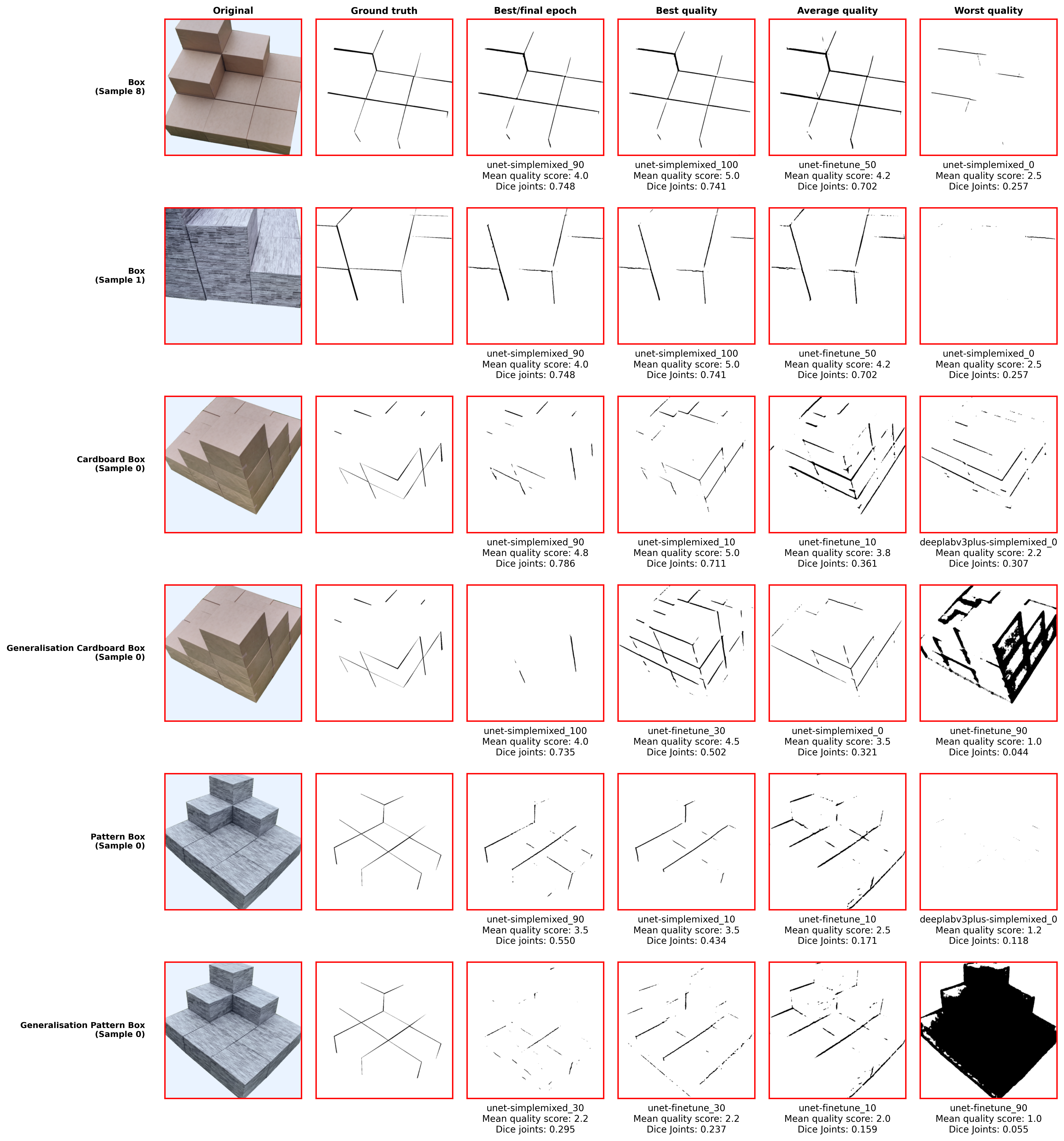}
\caption{Box-domain experiments: predicted masks at best/final epoch, best, average, and worst mean quality. Best epoch is shown for Finetune strategy whereas final epoch is shown for SimpleMixed strategy.}
\label{fig_best_worst_box}
\end{figure}

\begin{figure}[H]
\centering
\includegraphics[width=\linewidth]{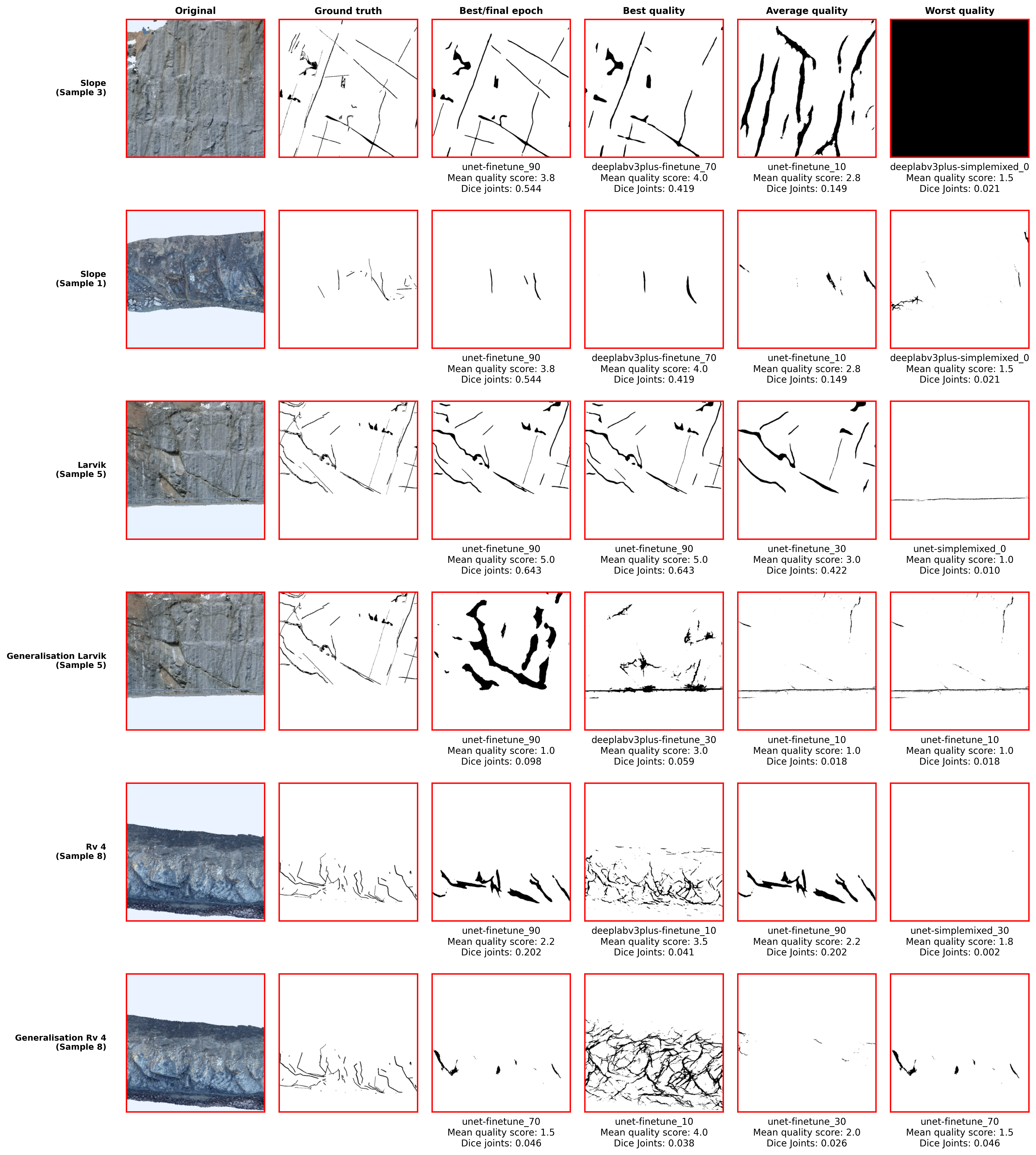}
\caption{Slope-domain experiments: predicted masks at best/final epoch, best, average, and worst mean quality. Best epoch is shown for Finetune strategy whereas final epoch is shown for SimpleMixed strategy.}
\label{fig_best_worst_slope}
\end{figure}

Representative masks (final/best, and highest/average/worst mean quality) per strategy are shown in Fig.~\ref{fig_best_worst_box} (box experiments) and Fig.~\ref{fig_best_worst_slope} (Slope experiments). Box exemplars illustrate improved continuity at best quality whereas fragmented joints and noise at worst. Slope exemplars show strong localisation for Larvik, but thicker or sparse predictions in Generalisation Larvik cases, Rv4 and its generalisation cases.

\paragraph{Box-domain experiments:}
In Box experiments, the highest mean quality scores are not always associated with the highest validation Dice (joints) (Fig.~\ref{fig_best_worst_box}). High-quality predictions are characterised by continuous, well-connected joint traces with realistic persistence, whereas some high-Dice cases exhibit fragmented detections or increased noise. Cardboard Box yields the strongest qualitative performance overall, notably SimpleMixed U-net can predict acceptable joint traces even with 0\% real data (Fig.~\ref{fig_qualitative_ratings_grid}), while Pattern Box and Generalisation Pattern Box predictions are more sensitive to texture-induced false positives (Fig.~\ref{fig_best_worst_box}).

SimpleMixed frequently matches or exceeds Finetune in mean quality score in Box experiments, particularly for U-net, consistent with the strong structural constraints imposed by synthetic training data (Fig.~\ref{fig_qualitative_ratings_grid}).

\paragraph{Slope-domain experiments:}
In Slope, particularly Larvik experiments, higher mean quality scores more often coincide with improved visual agreement with ground truth and are frequently observed alongside moderate to high validation Dice (joints) (Fig.~\ref{fig_best_worst_slope}). Larvik predictions commonly show good localisation and realistic joint persistence even when Dice values are not maximal. In contrast, Rv4 predictions are often sparse or overly thick, and generalisation cases frequently produce either limited detections or excessive false positives. Generalisation Larvik and Rv4 rarely achieve high mean quality scores, reflecting limited transferability across sites with differing geological conditions and image characteristics (Fig.~\ref{fig_qualitative_ratings_grid}).

In Larvik and its generalisation cases, Finetune more consistently yields higher mean quality scores than SimpleMixed (Fig.~\ref{fig_qualitative_ratings_grid}). However, no clear qualitative advantage of Finetune over SimpleMixed is observed for Rv4 and its generalisation cases. As highlighted in (Fig.~\ref{fig_qualitative_ratings_grid}), with Finetune strategies, the early epochs in Stage~2 (the fine-tuning phase) can predict better quality joint trace masks than the epoch with the best Dice (joints) in many cases. Exemplars of these are included in supplemetary material~\ref{sec:appendix_better_epochs}, which include generalisation experiments trained with a low proportion of real data.

\paragraph{Architectural comparison:}
Across most Box and Larvik experiments, U-net achieves mean quality scores comparable to or slightly higher than DeepLabv3+ (Fig.~\ref{fig_qualitative_ratings_grid}), whereas no consistent architectural differences are observed for Rv4 and generalisation cases, where qualitative performance is uniformly low.

\subsubsection{Correlation between quantitative and qualitative results}\label{sec:correlation_quantitative_qualitative}

\begin{figure}[H]
\centering
\includegraphics[width=\linewidth]{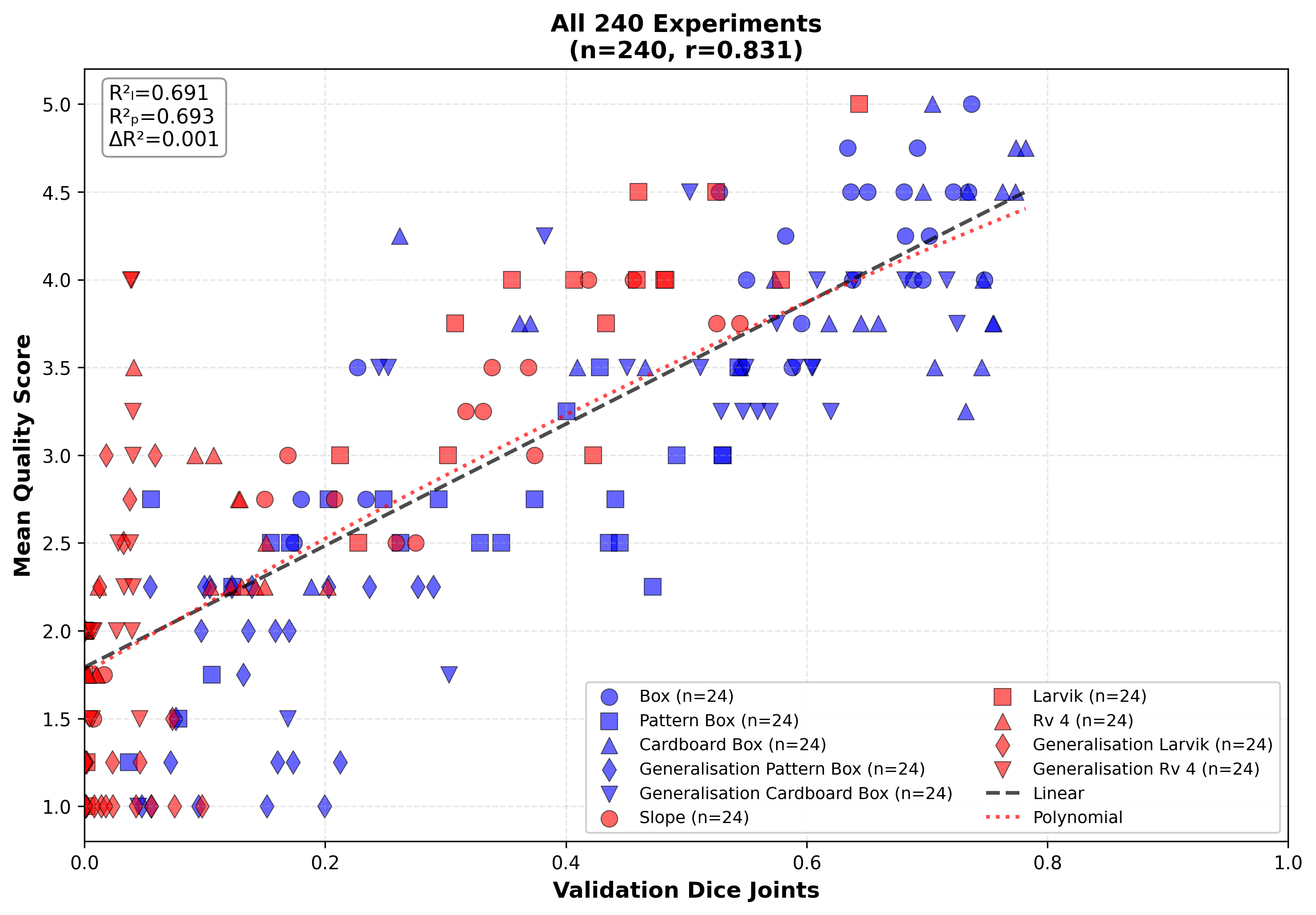}
\caption{Validation Dice (joints) vs mean quality score across all experiments. Linear and polynomial tendencies highlighted. Blue and red symbols represent box-domain and slope-domain experiments respectively. }
\label{fig_dice_vs_quality_all}
\end{figure}

Across all experiments, validation Dice (joints) and mean quality score show a strong positive correlation ($r = 0.831$) (Fig.~\ref{fig_dice_vs_quality_all}). However, this relationship is domain-dependent. For a given Dice value, Slope experiments (red symbols) in tend to achieve higher qualitative scores than Box experiments (blue symbols), indicating greater geological and engineering usability despite moderate overlap metrics.

\begin{figure}[H]
\centering
\includegraphics[width=0.7\linewidth]{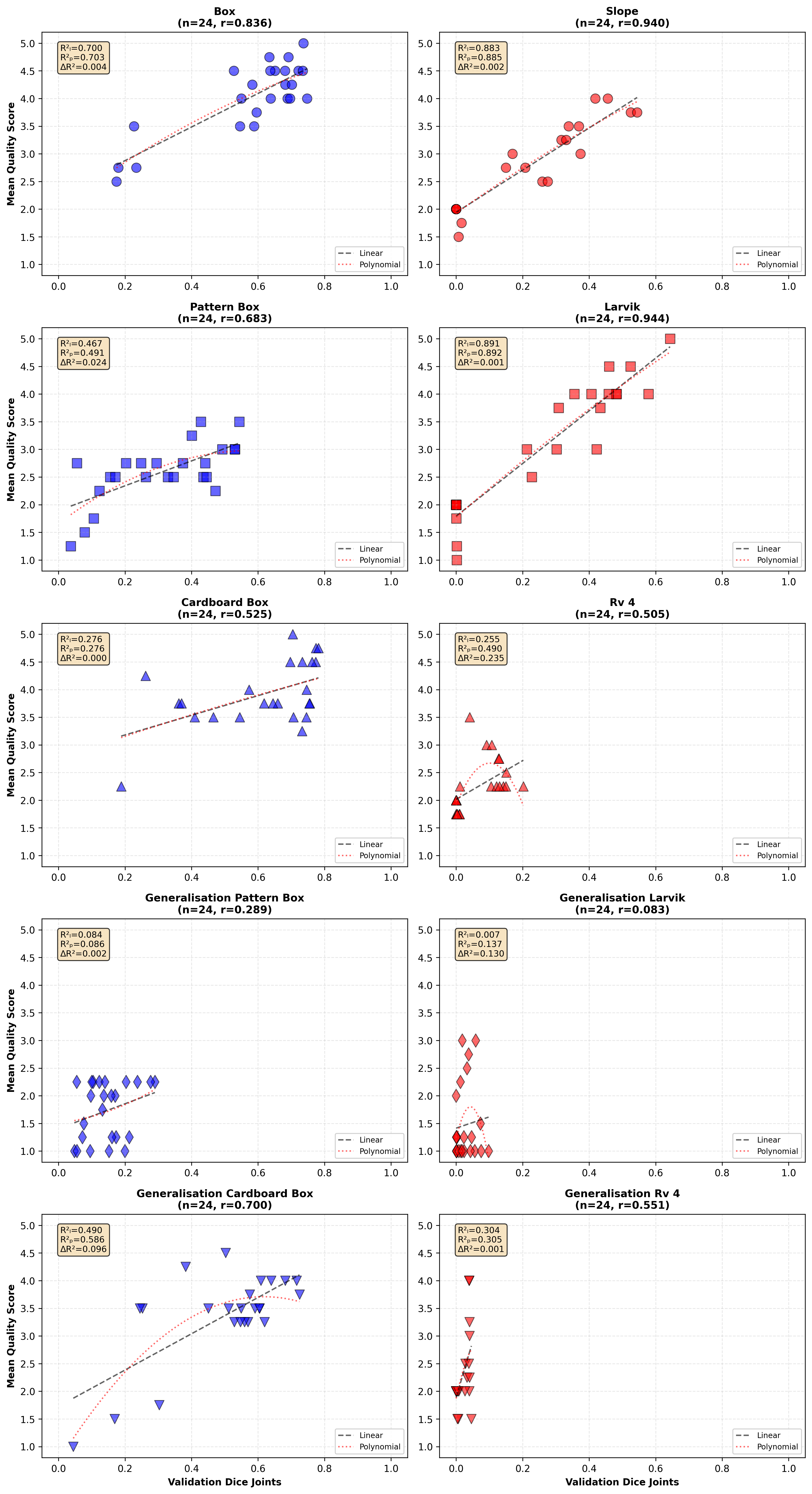}
\caption{Per-experiment correlation between validation Dice (joints) and mean quality. Linear and polynomial tendencies highlighted.  Blue and red symbols represent box-domain and slope-domain experiments respectively.}
\label{fig_dice_vs_quality_grid}
\end{figure}

At the per-experiment and per-strategy level, correlations range from moderate to strong ($r = 0.525$–$0.940$), with only marginal differences between linear and polynomial fits ($\Delta r^2 \le 0.024$) (Fig.~\ref{fig_dice_vs_quality_grid}). Non-linear tendencies are most evident in Rv4 and selected generalisation experiments, where polynomial fits yield higher explanatory power ($\Delta r^2 = 0.096$–$0.130$). Importantly, qualitative improvements are frequently realised through cleaner boundaries and more geologically plausible joint persistence without corresponding gains in validation Dice (joints). This effect is more pronounced in slope-domain experiments. A representative example is provided by the best-quality predicted masks for Rv4, which exhibit an extensive and geologically realistic joint trace network at an early epoch in Stage~2 (Fig.~\ref{fig_best_worst_slope}). This underscores the role of qualitative assessment as a necessary complement to quantitative metrics when evaluating joint-detection performance.

Overall, the results indicate that performance differences are primarily driven by domain characteristics and training strategy, with architectural effects being secondary and strongly domain-dependent. These observations motivate further discussion on the conditions under which synthetic data can support joint detection in real rock slope imagery and the limitations imposed by domain shift.

\section{Discussion}\label{sec:discussion}

This study presents a combined geological–\ac{ML} methodology for automated rock joint trace mapping, with the primary objective of developing a parameter-controlled and reproducible process for creating synthetic images of jointed rock masses for supervised \ac{ML}. The central contribution is the \ac{DFN}-based synthetic data workflow, which is designed to act as a foundation for joint trace detection by providing a structured geological prior that can be adapted using varying amounts of real data. \ac{ML} and evaluation are employed as means to verify whether this synthetic foundation can support supervised joint detection and generalise to real and previously unseen scenarios.This section evaluates the results against the three research questions stated in Section~\ref{sec:introduction}, as well as limitations, and outlook for future research.

\subsection{Synthetic data generation as a geological prior for machine learning}

The created synthetic \ac{DFN} dataset demonstrates that jointed rock images can be generated at field-relevant scales while preserving key geological characteristics of joint networks. The dominance of Y-nodes and connectivity indices above thresholds associated with purely random systems (Fig.~\ref{fig_network}) indicate that the generated trace networks reflect structured joint-set behaviour and termination logic. Similarly, the block volume and shape distributions across the 27 \acp{DFN} (Fig.~\ref{fig_block_volume_shape_distribution}) confirm that the parametric workflow spans a range of block geometries commonly encountered in rock engineering.

From a methodological perspective, the principal contribution is not realism alone, but also geological control. By explicitly parameterising block shape, joint set orientation and intensity, joint chronology, trace waviness, and visual thickness, the synthetic generator embeds geological assumptions directly into the training data. This provides the ML models with a structured geological prior that is difficult to obtain from limited real-world datasets, where mapping subjectivity, exposure conditions, and scale effects are inherently entangled.

\subsection{Effectiveness of synthetic data for supervised joint trace mapping}

The results show that synthetic data can support supervised joint trace detection, but the effect is domain-dependent. In the box-domain experiments, the SimpleMixed strategy demonstrates that synthetic data can effectively assist supervised joint trace learning with limited real data. In this domain, SimpleMixed reaches its best Dice  score (joints) and quality with little real data (0--10\%). In slope-domain experiments, Finetune strategy performs better than SimpleMixed. Pretraining on synthetic \ac{DFN} images followed by fine-tuning on real slope data improves recognisability and persistence, with a significant gain between 10\% and 50\% real data. The difference is that SimpleMixed is most effective in the box domain, whereas Finetune is most effective in the slope domain, due to differences in annotation quality and visual complexity between the two settings. Compared to the box-domain annotations, slope labels are noisier, less consistent in trace thickness and continuity, and more affected by occlusion, lighting, and scale effects. Under these conditions, SimpleMixed struggles because the synthetic and real label distributions differ strongly, whereas Finetune allows the model to first learn a stable structural prior from synthetic data and then adapt to the imperfect and site-specific real labels. SimpleMixed performs poorly at low real-data proportions in slopes and only improves at high proportions (70--100\%).

Zero-shot prediction from synthetic model remains limited overall but is feasible under favourable appearance conditions. In the Cardboard Box experiments, SimpleMixed U-net achieves the strongest qualitative zero-shot performance, predicting acceptable joint traces even without training with any real data. This contrasts with slope images, where dominant surface textures and illumination effects hinder transfer, indicating that zero-shot feasibility is primarily constrained by the appearance gap rather than joint-network geometry.

Across both domains, architectural choice dlimited influence on performance compared to the properties of the training data. U-net and DeepLabv3+ exhibit similar behaviour in general, with differences that are small relative to the effects of dataset composition, label quality, and domain shift. In well-controlled settings such as the box-domain experiments, multiple architectures already achieve near-optimal results, indicating that the primary bottleneck is not the algorithm but the dataset. In contrast, performance degrades across all architectures in slope-domain and generalisation experiments, reflecting the impact of label noise and appearance variability rather than architectural limitations. These results emphasise that the main contribution of this study lies in addressing data quality and generalisability through a parameter-controlled synthetic dataset, rather than in proposing new model architectures.

\subsection{Qualitative evaluation as a central methodological contribution}

This study challenges the assumption that higher pixel-wise accuracy does not necessarily implies better geological utility. In several cases, predictions with moderate Dice values exhibit clearer continuity, more realistic thickness, and fewer spurious detections than predictions achieving higher Dice scores. This highlights a fundamental limitation of relying solely on quantitative overlap metrics when the target features are thin, discontinuous, and subject to labelling uncertainty. Our key contribution is the introduction and systematic application of a structured qualitative evaluation framework tailored to rock joint trace interpretation. By explicitly rating geological recognisability, trace persistence, boundary localisation, and false positives, the evaluation aligns model assessment with how results are actually judged and used in rock engineering practice.

The strong overall correlation between validation Dice and mean quality score ($r = 0.831$) confirms that quantitative metrics remain informative. However, consistent deviations particularly in slope and generalisation experiments demonstrate that quantitative metrics alone is insufficient for selecting the most useful model state. An interesting observation is that qualitatively best predictions are often found at intermediate epochs, especially in the early epochs in Stage~2 (the fine-tuning stage) in Finetune strategy, rather than at the highest validation Dice (joints). The findings suggest that model selection and comparison for joint trace mapping should incorporate qualitative or geology-relevant criteria alongside conventional metrics, particularly when the ground truths can be biased, inaccurate, and incomplete.

A possible explanation for the misalignment between metrics and predicted joint mask quality is due to the training dynamics. In particular, the learning rate appears to affect the level of geometric detail and persistence. Smaller gradient descent steps may reduce sensitivity to local label noise and mitigate overfitting to a specific annotation detail level.  As a result, some models produce geologically plausible traces that extend beyond imperfect labels, leading to lower Dice scores despite higher geological usefulness. On the contrary, some models exhibit overfitting behaviour during Stage 2 (the fine-tuning stage) of Finetune strategy, resulting in visually degraded predictions at later epochs, such as fragmented traces or thickening of joints, as highlighted in Fig.  \ref{fig_qualitative_ratings_grid} and illustrated in supplementary material~\ref{sec:appendix_better_epochs}.

\subsection{Generalisation as a domain adaptation challenge}

Generalisation experiments across sites and textures yield low quantitative scores and reduced qualitative ratings, but these results should not be interpreted as a failure of the synthetic-data approach. Qualitative inspection shows that some low-scoring predictions remain geologically (supplementary material~\ref{sec:appendix_better_epochs}), suggesting that the model has learned relevant geometric and visual characteristics of rock joint traces. The results therefore motivate future work on domain adaptation techniques specifically tailored to rock engineering imagery, where appearance variation is large and labels are uncertain.

\subsection{Implications for production deployment}

The low Dice scores (joints) observed in slope-domain generalisation experiments indicate a substantial domain shift between training and test images, particularly across sites with differing surface textures, lighting conditions, and annotation styles. From a production perspective, this suggests that a single universal model is unlikely to be robust without adaptation. Instead, the most practical deployment strategy is to use the synthetic-pretrained model as a foundation and perform limited site-specific fine-tuning prior to operational use.

A realistic workflow therefore consists of maintaining a stable synthetic-trained baseline model and adapting it using a small number of labelled images from each new project. For example, in a new road cut, fine-tuning with up to a hundred representative images could be sufficient to calibrate the model before applying it to larger image collections. This approach is consistent with the slope-domain results, where fine-tuning significantly improves recognisability and persistence under noisy and site-specific labels.

As discussed earlier, training dynamics, particularly the learning rate, affect whether the model favours exploration of fine-scale joint details or exploitation of more persistent structural patterns. This observation suggests that a scale-aware joint detection framework can be achieved by selecting model states that reflect different learning behaviours, allowing a single trained model to support multiple engineering use cases without requiring fundamentally different architectures. For engineering geology applications, the required level of joint detail in the predicted traces depends on the intended use. Detailed joint representations are relevant for assessing highly fractured rock masses and support needs such as sprayed concrete, whereas simplified representations that emphasise persistent, block-forming discontinuities are sufficient for block or global stability assessments.

Although demonstrated here for rock slopes, the same synthetic data workflow can be extended to tunnel environments by adapting the geometry and appearance of the synthetic generator and fine-tuning on tunnel face images.

\subsection{Limitations and implications for engineering workflows}

This study is limited to 2D RGB imagery and does not incorporate explicit 3D geometric information from point clouds or depth maps, which have been shown in several related studies to improve joint trace discrimination, particularly in low-contrast or shadowed regions (e.g. \citealp{guo_geometry-_2019, guo_automatic_2022, mehrishal_new_2024}). The synthetic \ac{DFN} dataset further represents a simplified structural setting, excluding geological features such as bedding, foliation, folded discontinuities, and fault-related damage zones that may dominate joint patterns in some rock masses.

A further limitation relates to differences in image acquisition modalities across the real-world validation experiments. The real-world box experiment is based on camera photographs acquired from physical surfaces, whereas the real-world slope experiment is based on 2D rendered images derived from 3D point clouds. These modalities may differ in lighting consistency, surface texture representation, occlusion effects, and noise characteristics. In particular, rendered images may be sensitive to point-cloud density, surface reconstruction, and shading models, potentially leading to missing or altered visual information that affects joint trace detectability and limits direct comparability between experiments.

In addition, while the synthetic joint networks are created to preserve key statistical and topological characteristics of natural systems, the parametric \ac{DFN} process may also produce joints that are not strictly representative of natural fracture formation. In particular, some joint terminations, spacings, or intersections may be mechanically unrealistic or geologically unlikely when compared to site-specific joint genesis and stress history. Such artefacts are an inherent limitation of abstracted fracture network modelling and may introduce patterns that do not occur in nature. This reinforces the role of synthetic data in this study as a geological prior rather than a direct replica of real joint systems, and further motivates the need for adaptation using real data to constrain model behaviour toward site-specific realism.

Downstream post-processing steps, such as trace linking, vectorisation, network cleaning, and conversion into joint set statistics, are not implemented or evaluated here. These steps have been addressed in other studies (e.g. \citealp{chen_machine_2022, mehrishal_new_2024}), but their absence limits direct validation against engineering tasks such as block identification or kinematic analysis. The present results should be interpreted as assessing the quality of pixel-level trace detection rather than end-to-end engineering performance.

\subsection{Outlook}

Future work should extend the synthetic generator to improve appearance realism and cover a broader range of geological structures and, while maintaining parametric control that forms the basis of the current \ac{DFN}-based synthetic dataset workflow. 

Future work could combine the current \ac{DFN}-based synthetic data workflow with generative artificial intelligence and gaming technologies to improve visual realism. While \ac{DFN} modelling ensures geological control of joint network structure, generative and game-based approaches can better capture surface texture, lighting, and environmental variability, enabling synthetic datasets that remain structurally meaningful while more closely resembling real rock surfaces.

Beyond visual realism, an important open question is how the structural characteristics embedded in the synthetic data affect model transferability. Future work should investigate how specific joint network characteristics already embedded in the synthetic \ac{DFN} dataset influence model generalisability and adaptability to new geological settings. While the current synthetic generator controls various joint network parameters such as joint-set configuration, joint chronology and termination patterns, and network connectivity to ensure geological plausibility, their individual and combined effects on cross-site adaptation remain unclear. Systematic variation of these parameters in controlled synthetic experiments could help identify which aspects of joint network topology are most critical for robust adaptation.

Integrating post-processing and geological interpretation workflows will be essential to calibrate segmentation outputs against established joint mapping methods and to validate the combined geological–ML methodology in full engineering pipelines. Future studies should also explore skeletonisation or centreline extraction of predicted joint masks prior to metric calculation and downstream processing. Reducing predictions to one-pixel-wide representations could improve metric comparability by reducing sensitivity to trace thickness and minor spatial misalignments, while also supporting subsequent tasks such as trace linking, vectorisation, and network analysis.

The limitation of metric-based early stopping calls for the need for evaluation criteria that reflect geological properties such as trace directions, continuity and persistence. Future work should also address how qualitative judgement can be formalised in a systematic and reproducible manner. One promising direction is the development of structured qualitative evaluation protocols based on blind and randomised assessment of predicted joint masks by domain experts and non-experts. Such approaches could support the development of task-specific evaluation metrics that better align quantitative evaluation with geological usefulness, and provide a more robust basis for model comparison and selection in rock engineering applications.

In addition, systematic hyperparameter tuning should be incorporated to identify optimal training dynamics, including learning rate selection, prediction thresholds, and early-stopping criteria. Such optimisation would be important for reproducible and interpretable model comparisons.

\section{Conclusions}\label{sec:conclusions}

This study presents a novel geology-driven \ac{ML} methodology for automated rock joint trace mapping. Synthetic data generation, model training, and evaluation are co-designed to address data scarcity and class imbalance in real-world rock mass imagery. Geological knowledge is embedded through \ac{DFN}-based synthetic images. These images provide a strong structural prior for supervised learning. Synthetic data are not treated as a substitute for real observations. They are used to shape model behaviour and promote geologically meaningful predictions.

Synthetic jointed rock images are created at field-relevant scales and preserve key structural characteristics, including trace persistence, connectivity, and node-type distributions. These datasets are used to pretrain and train segmentation models using fine-tuning and mixed training strategies respectively. The results show that synthetic data can support supervised joint trace detection, but the prediction performance is domain-dependent and affected by the label quality. In the box-domain experiments, the SimpleMixed strategy achieves its best Dice (joints) scores and qualitative performance with only 0–10\% real data. In the slope-domain experiments, where labels are noisy, the Finetune strategy performs better, with clear improvements qualitatively between 10\% and 50\% real data. SimpleMixed shows limited benefit at low real-data proportions in the slope-domain and only improves at high proportions, typically above 70\%. The results show that synthetic data can act as an effective geological prior such that realistic joint trace prediction can be achieved in real images. Fully zero-shot transfer remains challenging unless synthetic images closely resemble real surface appearance, especially the texture.

Model performance was evaluated using quantitative metrics and qualitative assessment. Quantitative results show moderate to strong correlations across experiments. Qualitative assessment was therefore used as a decisive criterion to select the best-performing models. This assessment captures geological aspects that are not reflected in standard segmentation metrics when ground-truth annotations are biased, sparse, or internally inconsistent. The study demonstrates that expert-driven qualitative evaluation is not only complementary, but necessary for model selection in joint trace mapping until geologically meaningful quantitative metrics with statistical support become available.

The methodology was demonstrated on real-world rock slope and excavation datasets from Norwegian road infrastructure projects. Surface conditions, lighting, weathering, and annotation quality vary between sites. Synthetically trained models still recover geologically plausible joint trace maps. This capability is relevant for digital rock mass characterisation and preliminary joint mapping. It also supports downstream analyses that require consistent and reproducible joint information. Performance limitations are strongly linked to the quality and consistency of ground-truth data. Improved field data acquisition and annotation are therefore required, but our results utilising geology-driven synthetic data show promising potential to support supervise \ac{ML} when good quality real data is difficult to obtain.

The main contribution of this work is the demonstration that a combined geological and machine learning approach enables scalable and reproducible joint trace mapping under realistic data constraints. For rock engineering practitioners, the methodology supports automated joint mapping that complements manual interpretation. Geological reasoning remains central to the workflow. Future work should integrate geological post-processing of predicted traces. Statistically supported geological performance metrics should be developed. Domain adaptation and site-specific calibration should also be addressed. These steps are required to translate artificial intelligence-based joint detection into reliable tools for engineering design and decision-making.


\subsection*{Supplementary Information}

The supplementary materials provide external datasets, code repositories, and large collections of results that are impractical to include in the main manuscript. All supplementary materials are accessible online.

\begin{enumerate}[label=\arabic*]
    \item All data used for \ac{ML} training (rock mass images and corresponding labels), including dataset datasheets, available on Zenodo: \url{10.5281/zenodo.18078781}
    \item Real-world box dataset, including data acquisition and processing notes, raw images (RAW and JPEG), 3D point clouds (LAS), and processing reports from Agisoft Metashape, available on Zenodo: \url{10.5281/zenodo.18079090} \label{sec:suppl_realworld_box}
    \item Best performing models for rock joint segmentation, available on Zenodo: \url{10.5281/zenodo.18098009}
    \item Repository containing Grasshopper scripts for preparing synthetic datasets: \url{https://github.com/norwegian-geotechnical-institute/synthetic_rockmass_image_generation_for_ml}.
    \item Repository containing code for \ac{ML} training and evaluation pipelines: \url{https://github.com/tfha/syntetic_rock_joint_generation_for_ml}.
    \item Results of parametric modelling of the synthetic discrete fracture network dataset.\label{sec:appendix_dfn_results}
    \item Compiled plots showing the progression of training loss across experiments\label{sec:appendix_loss_plots}
    \item Compiled plots showing the progression of validation Dice score (joints)\label{sec:appendix_dice_joints_plots}
    \item Prediction masks from selected epochs for all 240 experiments
    \item Exemplars of experiments exhibiting higher-quality intermediate epochs\label{sec:appendix_better_epochs}
    \item Summary of best validation dice joints and quality scores of all 240 experiments
\end{enumerate}

\subsection*{Appendices}

The appendices provide additional methodological details, and documentation that complement the main text.

\begin{enumerate}[label=Appendix~\Alph*, leftmargin=*]
    \item Parametric modelling of the synthetic discrete fracture network dataset\label{sec:appendix_synthetic_dfn}
    \item REFORMS reproducibility checklist\label{sec:appendix_reforms}
\end{enumerate}

\subsection*{Acknowledgements}
We acknowledge the funding support from Research Council of Norway via STIPINST PhD grant (Grant No. 323307), Bever Control AS, and Bane NOR. We want to express our appreciation to Norwegian Geotechnical Institute for supplying the data for the Larvik and \ac{Rv4} case study. We thank Rocscience Inc, Maptek, WSP, and Itasca for providing the academic license for RocSlope3, PointStudio, FracMan, and DFN.lab respectively.

\subsection*{Author Contributions}

\textbf{Jessica Ka Yi Chiu}: Conceptualization, Methodology, Data Curation, Software, Investigation, Visualization, Formal Analysis, Writing – Original Draft, Writing – Review \& Editing.
\textbf{Tom Frode Hansen}: Conceptualization, Methodology, Software, Validation, Formal Analysis, Writing – Review \& Editing, Funding Acquisition.
\textbf{Eivind Magnus Paulsen}: Investigation, Data Curation, Software, Validation, Writing – Review \& Editing, Funding Acquisition.
\textbf{Ole Jakob Mengshoel}: Supervision, Writing – Original Draft, Writing – Review \& Editing.

\subsection*{Data and Code availability}
The datasets and code written and analysed during the current study are publicly
available through the repositories listed in the supplementary materials. The corresponding author may be contacted for questions regarding data access or reuse.

\section*{Statements and Declarations}

\subsection*{Use of Artificial Intelligence Tools}

The use of \acp{LLM} during software development is described in the Methodology section. All generated code was reviewed, tested, and modified where necessary by the authors, who take full responsibility for the final implementation and results.

\subsection*{Conflicts of Interest}
The authors declare that they have no known competing financial interests or personal relationships that could have appeared to influence the work reported in this paper.

\bibliography{references}% common bib file

\end{document}